 \documentclass[tablecaption=bottom,wcp]{jmlr} 

 \usepackage{tikz}
 \usetikzlibrary{arrows, calc}
\tikzset{
>=stealth',
help lines/.style={dashed, thick},
axis/.style={<=>},
important line/.style={thick},
connection/.style={thick, dotted},
}
 
 \usepackage{mathtools, lscape}

\usepackage{booktabs}
\usepackage[load-configurations=version-1]{siunitx} 

\DeclareMathOperator{\Ndash}{N-}

\theorembodyfont{\upshape}
\theoremheaderfont{\scshape}
\theorempostheader{:}
\theoremsep{\newline}

\jmlrproceedings{AABI 2019}{2nd Symposium on Advances in Approximate Bayesian Inference, 2019}

\title[Benchmarking Neural Linear]{Benchmarking the Neural Linear Model for Regression}

 \author{\Name{Sebastian W. Ober} \Email{swo25@cam.ac.uk}\\
 \addr University of Cambridge, Cambridge, UK
  \AND
 \Name{Carl Edward Rasmussen} \Email{cer54@cam.ac.uk}\\
  \addr University of Cambridge, Cambridge, UK
  \\
  \addr PROWLER.io, Cambridge, UK
 }

\begin{document}

\maketitle

\begin{abstract}
The neural linear model is a simple adaptive Bayesian linear regression method that has recently been used in a number of problems ranging from Bayesian optimization to reinforcement learning. Despite its apparent successes in these settings, to the best of our knowledge there has been no systematic exploration of its capabilities on simple regression tasks. In this work we characterize these on the UCI datasets, a popular benchmark for Bayesian regression models, as well as on the recently introduced UCI ``gap'' datasets, which are better tests of out-of-distribution uncertainty. We demonstrate that the neural linear model is a simple method that shows generally good performance on these tasks, but at the cost of requiring good hyperparameter tuning.
\end{abstract}


\section{Introduction}
Despite the recent successes that neural networks have shown in an impressive range of tasks, they tend to be overconfident in their predictions \citep{guo2017calibration}. Bayesian neural networks (BNNs; \citet{neal1995bayesian}) attempt to address this by providing a principled framework for uncertainty estimation in predictions. However, inference in BNNs is intractable to compute, requiring approximate inference techniques. Of these, Monte Carlo methods and variational methods, including Monte Carlo dropout (MCD) \citep{gal2016dropout}, are popular; however, the former are difficult to tune, and the latter are often limited in their expressiveness \citep{foong2019between, yao2019quality, foong2019pathologies}.

The neural linear model represents a compromise between tractability and expressiveness for BNNs in regression settings: instead of attempting to perform approximate inference over the entire set of weights, it performs exact inference on only the last layer, where prediction can be done in closed form. It has recently been used in active learning \citep{pinsler2019bayesian}, Bayesian optimization \citep{snoek2015scalable}, reinforcement learning \citep{riquelme2018deep}, and AutoML \citep{zhou2019adaptive}, among others; however, to the best of our knowledge, there has been no systematic attempt to benchmark the model in the simple regression setting. In this work we do so, first demonstrating the model on a toy example, followed by experiments on the popular UCI datasets (as in \citet{hernandez2015probabilistic}) and the recent UCI gap datasets from \citet{foong2019between}, who identified (along with \citet{yao2019quality}) well-calibrated `in-between' uncertainty as a desirable feature of BNNs.

\section{Methods}
\begin{figure}[htbp]
\floatconts
  {fig:slice_plots}
  {\caption{Predictive distributions for the toy problem. Each shaded region represents one predictive standard deviation.}}
  {\includegraphics[width=\linewidth]{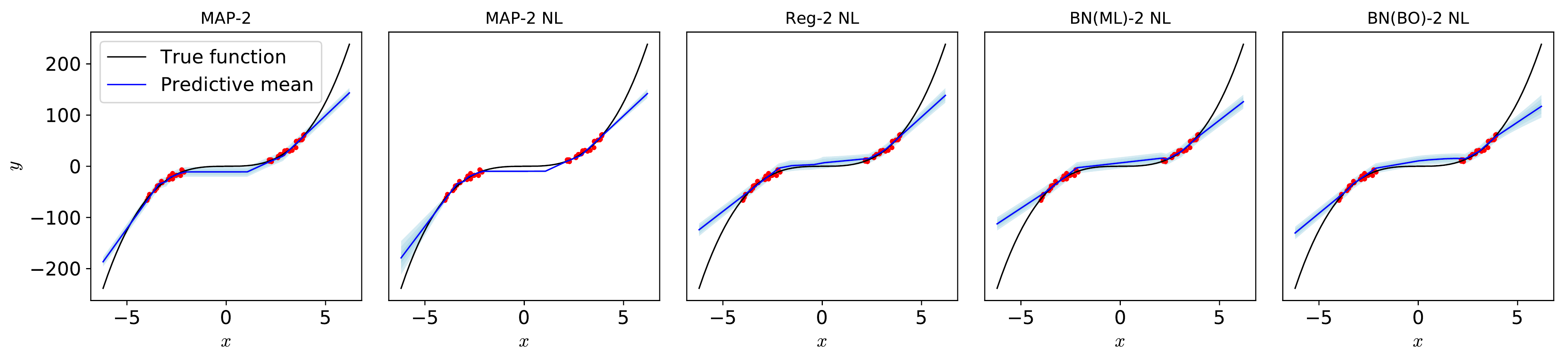}}
\end{figure}
In this section, we briefly describe the different models we train in this work, which are variations of the neural linear (NL) model, in which a neural network extracts features from the input to be used as basis functions for Bayesian linear regression. The central issue in the neural linear model is how to train the network: in this work, we provide three different models, with a total of four different training methods. For a more complete mathematical description of the models, refer to Appendix \ref{apd:NL}; we summarize the models in Appendix \ref{apd:model_summ}.

\paragraph{MAP NL} Following the work of \citet{snoek2015scalable}, we can first train the neural network using maximum a posteriori (MAP) estimation. After this training phase, the outputs of the last hidden layer of the network are used as the features for Bayesian linear regression. To reduce overfitting, the noise variance and prior variance (for the Bayesian linear regression) are subsequently marginalized out by slice sampling \citep{neal2003slice} according to the tractable marginal likelihood, using uniform priors. We refer to this model as the \textit{maximum a posteriori neural linear model} (which we abbreviate as {MAP-$L$ NL}, where $L$ is the number of hidden layers in the network). We tune the hyperparameters for the MAP estimation via Bayesian optimization \citep{snoek2012practical}.

\paragraph{Regularized NL} The MAP NL model's basis functions are learned independently of the final model's predictions. This is an issue for uncertainty quantification, as MAP training has no incentive to learn features useful for providing uncertainty in out-of-distribution areas. To address this issue, we propose to learn the features by optimizing the (tractable) marginal likelihood with respect to the network weights (previous to the output layer), treating them as hyperparameters of the model in an approach analogous to hyperparameter optimization in Gaussian process (GP) regression \citep{rasmussen2006gaussian}. However, unlike in GP regression, the per-iteration computational cost of this method is linear in the size of the data. We additionally regularize the weights to reduce overfitting, resulting in a model we call \textit{regularized neural linear} (which we abbreviate as \textit{Reg-$L$ NL}). As in the MAP NL model, we marginalize out the noise and prior variances via slice sampling. We tune the regularization and other hyperparameters via Bayesian optimization.

\paragraph{Bayesian noise NL} Instead of using slice sampling for the noise variance, we can place a normal-inverse-gamma (N-$\Gamma^{-1}$) prior on the weights and noise variance. This formulation is still tractable, and integrates the marginalization of the noise variance into the model itself, rather than having it implemented after the features are learned. Additionally, the N-$\Gamma^{-1}$ prior can act as a regularizer, meaning that we can avoid using Bayesian optimization to tune the prior parameters by jointly optimizing the marginal likelihood over all hyperparameters. However, this risks overfitting. Therefore, we consider training this model, which we call the \textit{Bayesian noise (BN) neural linear} model, both by maximizing the marginal likelihood for all parameters (including prior parameters), and by tuning the prior parameters with Bayesian optimization. We abbreviate the first as \textit{BN(ML)-$L$ NL} and the second as \textit{BN(BO)-$L$ NL}. Finally, in both cases we slice sample the remaining (non-weight) hyperparameters.

\section{Experiments} \label{sec:Exp}
\begin{figure}[htbp]
\floatconts
  {fig:UCI_LLs}
  {\caption{Average test log likelihoods (nats) for the UCI datasets}}
  {\includegraphics[width=\linewidth]{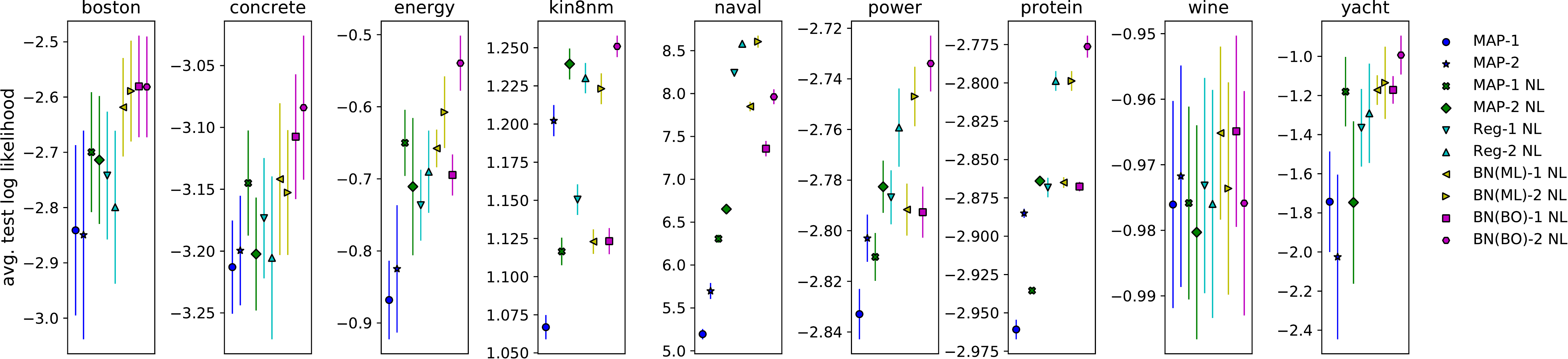}}
\end{figure}
We compare these models on a toy problem, the UCI datasets, and the UCI ``gap'' datasets \citep{foong2019between}. In all experiments, we consider 1- and 2-layer ReLU fully-connected networks with 50 hidden units in each layer (except for the toy problem, where we only consider 2-layer networks). We also provide results for simple MAP inference as a baseline. For experimental details, refer to Appendix \ref{apd:details}. We provide additional experimental results, including detailed statistical comparisons of the models, in Appendix \ref{apd:add_results}.

\paragraph{Toy problem} We construct a synthetic 1-D dataset comprising 100 train and 100 test pairs $(x, y)$, where $x$ is sampled i.i.d.~in the range $[-4, -2]\cup[2, 4]$ and $y$ is generated as $y= x^3 + \epsilon$, $\epsilon\sim \mathcal{N}(0, 9)$. This follows the example from \citet{hernandez2015probabilistic}, with the exception of the ``gap'' added in the range for $x$, which was motivated by \citet{foong2019between} and \citet{yao2019quality}. We plot predictive distributions for each model in \figureref{fig:slice_plots}. Somewhat surprisingly, the MAP-2 NL model seems to struggle more than MAP with uncertainty in the gap, while having better uncertainty quantification at the edges. Of the marginal likelihood-based methods, the BN(BO)-2 NL model qualitatively seems to perform the best.

\paragraph{UCI datasets} We next provide results on the UCI datasets in \citet{hernandez2015probabilistic} (omitting the `year' dataset due to its size), a popular benchmark for Bayesian regression models in recent years. We report average test log likelihoods and RMSEs for all the models in Appendix \ref{apd:full_results}, for both 1- and 2-layer architectures. We visualize average test log likelihoods for the models in \figureref{fig:UCI_LLs}; we tabulate the log likelihoods and RMSEs in \tableref{tab:UCI_LL,tab:UCI_RMSE} in Appendix \ref{apd:full_results}, respectively. 

From the figure and tables, we see that the BN(ML)-2 NL and BN(BO)-2 NL models have the best performance on these metrics, with reasonable log likelihoods and RMSEs compared to those in the literature for other BNN-based methods \citep{hernandez2015probabilistic, gal2016dropout, bui2016deep, tomczak2018neural}. In fact, these neural linear methods tend to achieve state-of-the-art or near state-of-the-art neural network performance on the `energy' and `naval' datasets. While the performance of the Reg-$L$ NL model is decent, it performs worse than the BN-$L$ NL models, showing the advantage of a Bayesian treatment of the noise variance.

\paragraph{UCI gap datasets} Finally, we provide results on the UCI ``gap'' datasets proposed by \citet{foong2019between}, which consists of training and testing splits that artificially contain gaps in the training set, ensuring that the model will only succeed if it can represent uncertainty in-between gaps in the data. We again visualize test log likelihoods in \figureref{fig:UCI_gap_LLs} while tabulating log likelihoods and RMSEs in \tableref{tab:UCI_gap_LL,tab:UCI_gap_RMSE} in Appendix \ref{apd:full_results}.

\begin{figure}[htbp]
\floatconts
  {fig:UCI_gap_LLs}
  {\caption{Average test log likelihoods (nats) for the UCI gap datasets}}
  {\includegraphics[width=\linewidth]{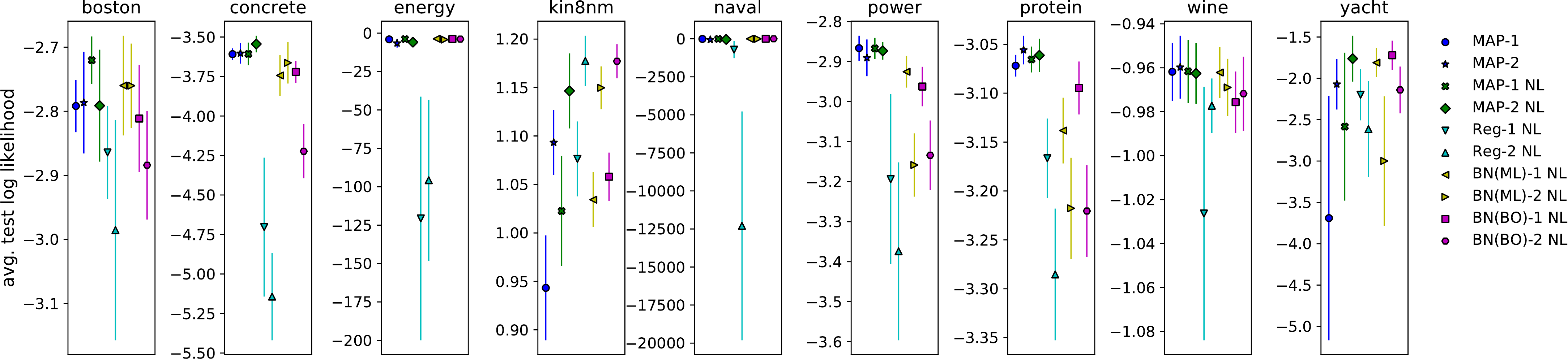}}
\end{figure}

Our results on the MAP-based models in \figureref{fig:UCI_gap_LLs} echo those of \citet{foong2019between}, showing catastrophic failure to express in-between uncertainty for some datasets (particularly `energy' and `naval'). Somewhat surprisingly, the Reg-$L$ NL models perform the worst of all the models. However, the BN NL models do not seem to fail catastrophically, with the BN(BO)-2 NL model having by far the best performance.

\section{The Effect of Hyperparameter Tuning}
\begin{figure}[htbp]
\floatconts
  {fig:single_diffs_UCI}
  {\caption{Differences in average test log likelihoods (nats) and RMSEs given by hyperparameter tuning for the UCI datasets}}
  {\includegraphics[width=\linewidth]{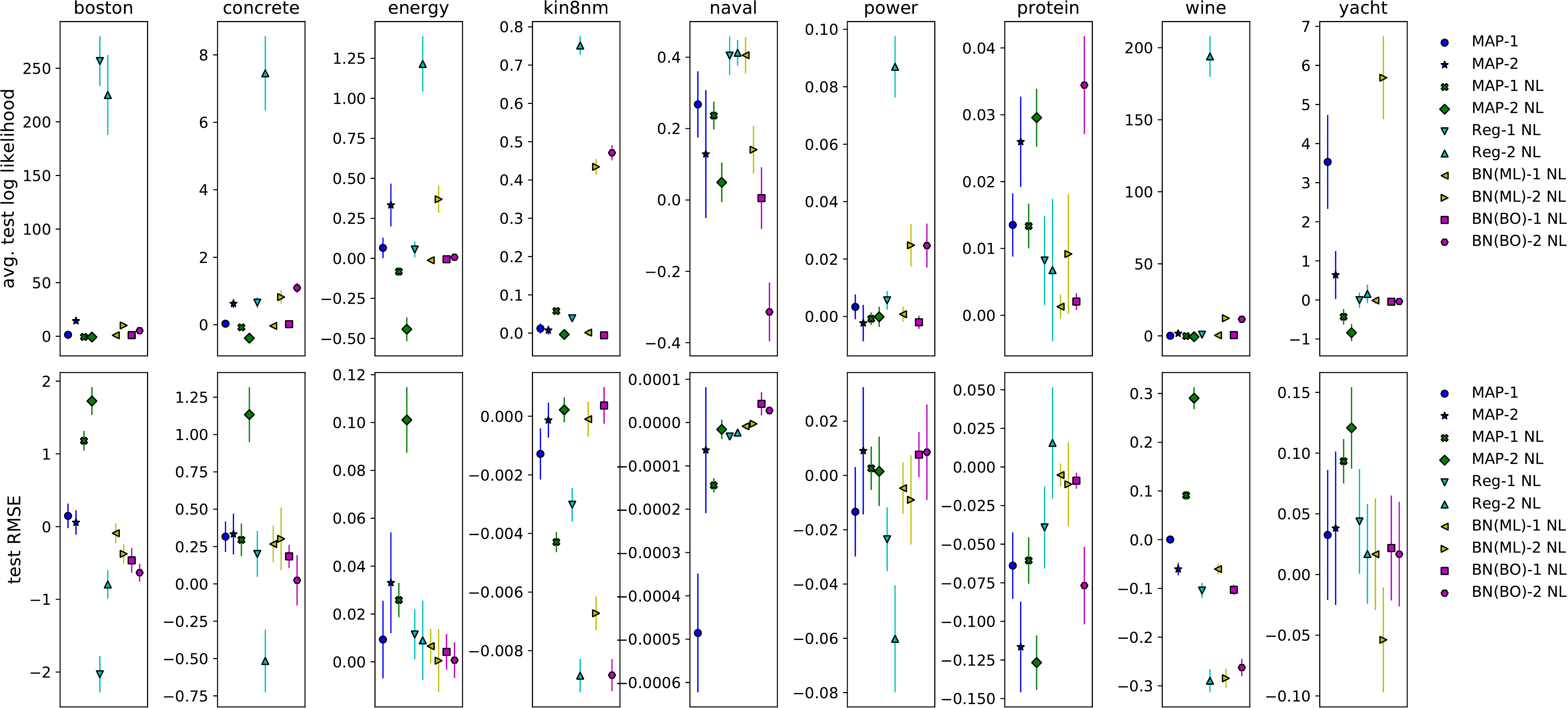}}
\end{figure}

In all of these models we used some form of hyperparameter tuning (Bayesian optimization for all models except the BN(ML)-$L$ NL models, where we used a grid search) to obtain the results shown. However, for the practitioner, performing an oftentimes costly hyperparameter search is not desirable, particularly where one of the main motivations for using the model is its simplicity, as in this case. We therefore investigate the effect of the hyperparameter tuning on the models' performance.

\figureref{fig:single_diffs_UCI} shows the difference in average test log likelihoods and test RMSEs between the tuned models and models whose hyperparameters were set to ``reasonable'' values that a practitioner might choose by intuition (see Appendix \ref{apd:hyp_tune} for details) for the UCI datasets. We observe that for each of the two-layer models there exists at least one dataset where the performance in terms of test log likelihood is \textit{significantly} worsened by omitting hyperparameter tuning. The performance difference for RMSEs is not as drastic, although it still exists. In Appendix \ref{apd:hyp_tune} we show that these results extend to the UCI gap datasets and that the difference in performance is statistically significant for nearly all models across both the UCI and UCI gap datasets, for both log likelihood and RMSE performance. Finally, in Appendix \ref{apd:Ms} we show that mean field variational inference (MFVI) \citep{graves2011practical, hinton1993keeping, blundell2015weight} and MCD can still obtain reasonable, although not state-of-the-art, performance on the UCI datasets without hyperparameter tuning: in many cases the performance is even competitive with the tuned NL models. However, these suffer from the pathologies identified in \citet{foong2019between, yao2019quality, foong2019pathologies} on the gap datasets.

\section{Conclusions \& Future Work}

We have shown benchmark results for different variants of the neural linear model in the regression setting. Our results show that the successes these models have seen in other areas such as reinforcement and active learning are not unmerited, with the models achieving generally good performance despite their simplicity. Furthermore, they are not as susceptible to the the inability to express gap uncertainty as MFVI or MCD. However, we have shown that to obtain reasonable performance extensive hyperparameter tuning is often required, unlike MFVI or MCD. Finally, our work suggests that exact inference on a subset of parameters can perform better than approximate inference on the entire set, at least for BNNs. We believe this broader issue is worthy of further investigation.

\section*{Acknowledgments}
We thank Andrew Y.K.~Foong, David R.~Burt, Ross M.~Clarke, and Thang D.~Bui for helpful discussions. We additionally thank Andrew Y.K.~Foong for providing the ``gap'' splits. SWO acknowledges the support of the Gates Cambridge Trust for his doctoral studies.

\bibliography{jmlr-sample}

\newpage
\appendix

\section{The Neural Linear Model} \label{apd:NL}
The neural linear model uses a neural network to parameterize basis functions for Bayesian linear regression by treating the output weights and bias of the network probabilistically, while treating the rest of the network's parameters $\theta$ as hyperparameters. This can be used as an approximation to full Bayesian inference of the neural network's parameters, with the main advantage being that this simplified case is tractable (assuming Gaussian prior and likelihood). Given the fact that there are significant redundancies in the weight-space posterior for BNNs, this tradeoff may not be a completely unreasonable approximation.

We now describe the model mathematically. Let $\mathcal{D} \coloneqq (X, Y) = \{(x_n, y_n)\}_{n=1}^N$, where $(x_n, y_n) \in \mathbb{R}^d\times\mathbb{R}$, be the training data, and let 
\begin{align*}
    \phi_{\theta}(x_n) &= [\phi_{\theta, 1}(x_n), \dots, \phi_{\theta, N_L}(x_n)]^T 
\end{align*} 
represent the outputs (post-activations) of the last hidden layer of the neural network, which will be parameterized by all the weights and biases up to the last layer, $\theta$. We then define a weight vector $w\in\mathbb{R}^{M}=\mathbb{R}^{N_L + 1}$ (this includes a bias term, augmenting $\phi_{\theta}(x)$ with a 1). If we define a design matrix $\Phi_{\theta} = [\phi_{\theta}(x_1), \dots, \phi_{\theta}(x_N)]^T$, we can then define our model as
\begin{align*}
    Y = \Phi_{\theta}w + \epsilon, \;\;\; \epsilon\sim\mathcal{N}(0, \sigma^2),
\end{align*}
where we treat $Y$ as a column vector of the $y_n$. Given an appropriate $\theta$, Bayesian inference of the weights $w$ is straightforward: given a prior $p(w) = \mathcal{N}(w;\, 0,\, \alpha\mathrm{I}_M)$ on the weights, the posterior is given by
\begin{align*}
    p(w|\mathcal{D}, \sigma^2) &= \mathcal{N}(w|w_N, \mathrm{V}_N)\propto \mathcal{N}(w;\, 0,\, \alpha\mathrm{I}_M)\,\mathcal{N}(Y;\,\Phi_{\theta}w,\, \sigma^2) , \\
    w_N &= \frac{1}{\sigma^2}\mathrm{V}_N\Phi_{\theta}^TY, \\
    \mathrm{V}_N^{-1} &= \frac{1}{\alpha}\mathrm{I}_M + \frac{1}{\sigma^2}\Phi_{\theta}^T\Phi_{\theta}.
\end{align*}
The posterior predictive for a test input $x_*$ is then given by
\begin{align*}
    p(y_*|x_*, \mathcal{D}, \sigma^2) = \mathcal{N}\left(y_*; w_N^T\phi_{\theta}(x_*),\, \sigma^2 + \phi_{\theta}(x_*)^T\mathrm{V}_N\phi_{\theta}(x_*)\right).
\end{align*}
It now remains to be determined how to learn $\theta$. 

\subsection{MAP Neural Linear}
\label{sec:MAP_NL}
As described in \citet{snoek2015scalable}, we can learn $\theta$ by simply setting it to the values of the corresponding weights and biases in a maximum a posteriori (MAP)-trained network, maximizing the objective
\[\mathcal{L}_{MAP}(\theta_{Full}) = \log{\mathcal{N}(Y;\,\Phi_{\theta}w,\, \sigma^2)} - \gamma\lVert\theta_{Full}\rVert_2^2,\]
with respect to $\theta_{Full}$ and $\sigma^2$, where $\theta_{Full}$ represents the parameters of the full network (which includes the output weights and bias), and $\gamma$ is a regularization parameter. As in \citet{snoek2015scalable}, once we have obtained $\theta$ from $\theta_{Full}$, we use can use Bayesian linear regression as outlined above. However, the question of setting $\alpha$ still remains. To address this, we marginalize $\alpha$ and $\sigma^2$ out by slice sampling them according to the log marginal likelihood of the data:
\begin{align*}
    \mathcal{L}_{\theta, \alpha, \sigma^2}(\mathcal{D}) &= \log{p(Y|X, \sigma^2)} = \log{\int p(Y|X,w,\sigma^2)\,p(w)\,\mathrm{d}w} \\
    &= -\frac{M}{2}\log{\alpha} - \frac{N}{2}\log{\sigma^2}-\frac{1}{2\sigma^2}||Y-\Phi_{\theta}w_N||^2_2 -\frac{1}{2\alpha}w_N^Tw_N \\
    &\;\;\;\;-\frac{1}{2}\log{|\mathrm{V}_N|} - \frac{N}{2}\log{2\pi}.
\end{align*}
In order to learn a suitable value of $\gamma$, along with learning rates and number of epochs, we use Bayesian optimization. For a complete description of the experimental details, see Appendix \ref{apd:details}.

One key disadvantage of this approach is that it separates the feature learning from prediction: in particular, there is no reason for the network to learn features relevant for out-of-distribution prediction, particularly when it comes to uncertainty estimates.

\subsection{Regularized Neural Linear}
\label{sec:Reg_NL}
From a Bayesian perspective, the neural linear model can be interpreted as a Gaussian process model with a covariance kernel determined by a finite number of basis functions $\phi_{\theta, i}$ with hyperparameters $\theta$. Therefore, as in Gaussian process regression, we propose to maximize the log marginal likelihood of the data, $\mathcal{L}_{\theta, \alpha, \sigma^2}(\mathcal{D})$, with respect to $\theta$ and $\sigma$ as the hyperparameters of the model for an empirical Bayes approach. 

Note that the computational complexity of this expression is $\mathcal{O}(N+M^3)$, as opposed to the $\mathcal{O}(N^3)$ cost typically seen in GP regression. This is because we are able to apply the Woodbury identity to obtain the determinant in terms of $\mathrm{V}_N$, which is $M\times M$, due to the fact that there is a finite number of basis functions. Since we typically have that $N\gg M$, this results in significant computational savings.

One issue with this Type-2 maximum likelihood approach is that it will tend to overfit to the training data due to the large number of hyperparameters $\theta$. As a result, the noise variance $\sigma^2$ will tend to be pushed towards zero. One way of addressing this is by introducing a regularization scheme. There are many potential regularization schemes that could be introduced: we could regularize $\theta$, $\alpha$, or $\sigma$ individually, or using any combination of the three. We found empirically that of these, simply regularizing $\theta$ alone via $L^2$ regularization seemed the most promising approach. This results in a Type-2 MAP approach wherein we maximize
\begin{align*}
    \hat{\mathcal{L}}_{\theta, \alpha, \sigma^2}(\mathcal{D}) = \mathcal{L}_{\theta, \alpha, \sigma^2}(\mathcal{D}) - \gamma_W\lVert\theta_W\rVert_2^2 - \gamma_b\lVert\theta_b\rVert_2^2,
\end{align*}
where we have divided $\theta$ into weights $\theta_W$ and biases $\theta_b$ and introduced regularization hyperparameters $\gamma_W$ and $\gamma_b$. 

\subsection{Bayesian Noise Neural Linear}
\label{sec:BN_NL}
An alternative to regularization would be to treat the noise variance in a Bayesian manner by integrating it out. Fortunately, for Bayesian linear regression this is still tractable with the use of a normal-inverse-gamma prior on the outputs weights and parameters
\begin{align*}
    p(w, \sigma^2) &= \Ndash\Gamma^{-1}(w, \sigma^2;\,0,\,\alpha\mathrm{I}_M,\,a_0,\,b_0) \\
    &= \mathcal{N}(w;\,0, \,\sigma^2\alpha\mathrm{I}_M)\Gamma^{-1}(\sigma^2;\,a_0,\, b_0).
\end{align*}
The posterior has the form
\begin{align*}
    p(w, \sigma^2|\mathcal{D}) &= \Ndash\Gamma^{-1}(w, \sigma^2; w_N, \mathrm{V}_N, a_N, b_N), \\
    w_N &= \mathrm{V}_N\Phi_{\theta}^TY, \\
    \mathrm{V}_N^{-1} &= \frac{1}{\alpha}\textrm{I}_M + \Phi_{\theta}^T\Phi_{\theta}, \\
    a_N &= a_0 + N/2, \\
    b_N &= b_0 + \frac{1}{2}(Y^TY - w_N^TV_N^{-1}w_N),
\end{align*}
with posterior predictive
\begin{align*}
    p(y_*|x_*, \mathcal{D}) = \mathcal{T}\left(y_*;\,w_N^T\phi_{\theta}(x_*),\, \frac{b_N}{a_N}(\mathrm{I}_M+\phi_{\theta}(x_*)^T\mathrm{V}_N\phi_{\theta}(x_*)),\, 2a_N\right),
\end{align*}
where $\mathcal{T}(\,\cdot\:;\,\mu,\,\Sigma, \,\nu)$ is a Student's t-distribution with mean $\mu$, scale $\Sigma$, and degrees of freedom $\nu$.  As before, we train the network using empirical Bayes, where the marginal likelihood is given by
\begin{align*}
    p(Y|X) = \mathcal{T}\left(Y;\,0,\,\frac{b_0}{a_0}(\mathrm{I}_M + \alpha\Phi_{\theta}\Phi_{\theta}^T),\, 2a_0\right).
\end{align*}
Note that by using the Woodbury identity it is possible to compute this in $\mathcal{O}(N+ M^3)$ computational cost as before.

\section{Experimental Details} \label{apd:details}
All neural networks tested were ReLU networks with one or two 50-unit hidden layers. When using a validation set, we set its size to be one fifth of the size of the training set, except for the toy example, where we used half the training set. We now describe the experimental setup for each model we used.

\paragraph{MAP} For the MAP baseline, we select a batch size of 32. We subsequently use Bayesian optimization (see section \ref{apd:BO} for a description of the Bayesian optimization algorithm we use) to optimize four hyperparameters using validation log likelihood: the regularization parameter $\gamma$, a learning rate for the weights, a learning rate for the noise variance, and the number of epochs. The regularization parameter is allowed to vary within the range corresponding to a log prior variance between -5 and 5. The learning rates are also optimized in log space in the range $[\log{\textrm{1e-4}}, \log{1e-2}]$. Finally, the number of epochs is set to vary between zero and the number required to obtain at least 10000 gradient steps (the number of epochs will thus vary with the size of the dataset given a constant batch size). We initialize the regularization parameter to 0.5, the learning rates at 1e-3, the noise variance at $e^{-3}$, and the number of epochs at the maximum value. The network itself is optimized using ADAM \citep{kingma2014adam}.

\paragraph{MAP NL} For the MAP neural linear model, we take the above optimal MAP network and obtain 200 slice samples of $\alpha_W$ (the output weight prior variance), $\alpha_b$ (the output bias prior variance), and $\sigma^2$ for Bayesian linear regression. We initialize $\alpha_W = 1/50$ and $\alpha_b = 1$, to match the scaling used in \citet{neal1995bayesian}.

\paragraph{Regularized NL} For the regularized NL model, there are five hyperparameters which we tune via Bayesian optimization: $\gamma_W$, $\gamma_b$, a learning rate for $\theta$, a learning rate for $\sigma^2$, and the number of epochs. We allow $\gamma_W$ and $\gamma_b$ to vary within a range of log prior variances between -10 and 10, and the number of epochs to be in the range of $[0, 5000]$ (since each epoch corresponds to one gradient step). The ranges for the other parameters remain the same. We initialize $\gamma_W$ and $\gamma_b$ to 1, and the remaining parameters the same way as in the MAP model. We again initialize $\alpha_W = 1/50$ and $\alpha_b = 1$. As before, we use 200 slice samples to marginalize out $\sigma^2$, $\alpha_W$, and $\alpha_b$ after the Bayesian optimization was completed.

\paragraph{Bayesian noise NL (ML)} Here we optimize the parameters $\theta$, $a_0$, $b_0$, $\alpha_W$, and $\alpha_b$ directly and jointly via the log marginal likelihood. We employ early stopping by tracking the validation log likelihood up to 5000 epochs, and also maximize the validation log likelihood over a grid of 10 learning rates ranging logarithmically from $\log{\textrm{1e-4}}$ to $\log{\textrm{1e-2}}$. We also initialize $a_0=b_0=1$ and $\alpha_W=\alpha_b=1$. Finally, we use slice sampling to obtain 200 samples to marginalize out these hyperparameters.

\paragraph{Bayesian noise NL (BO)} Instead of optimizing over the hyperparameters jointly as in the BN(ML) model, we keep all except $\theta$ fixed over each iteration of Bayesian optimization. We retain the same initializations, and allow the following ranges for the hyperparameters: $a_0\in[0, 20]$, $b_0\in [0, 10]$, $\log{\alpha_W} \in [-10, 10]$, $\log{\alpha_b}\in [-10, 10]$, with the ranges for the learning rate and number of epochs being the same as before. We retain the same initializations as before as well. The slice sampling also remains the same.

\subsection{Bayesian Optimization}
\label{apd:BO}
Here we describe the Bayesian optimization algorithm that we used throughout. In each case we attempt to maximize the validation log likelihood. We largely follow the formulation set out in \citet{snoek2012practical}. We use a Gaussian process with a Mat\'{e}rn-5/2 kernel with the model hyperparameters as inputs and the validation log likelihoods as outputs (normalizing the inputs and outputs). We first learn the kernel hyperparameters (including a noise variance) by maximizing the marginal likelihood of the GP, using 5000 iterations of ADAM \citep{kingma2014adam} with a learning rate of 1e-2. We then obtain 20 slice samples of the GP hyperparameters, before using the expected improvement acquisition function to find the next set of network hyperparameters to test. In total, we use 50 iterations of Bayesian optimization for each model, initialized with 10 iterations of random search.

\section{Summary of the Models} \label{apd:model_summ}
In \tableref{tab:model_summ}, we provide a summary of the models we use, describing which parameters are optimized and how (we exclude learning rates and the number of epochs from this table), and which are then slice sampled.

\begin{table}
\small
\floatconts
  {tab:model_summ}%
  {\caption{Summary of the models presented. The first column lists the model; the second shows the optimization objective, while the third shows which parameters were optimized using this objective. Meanwhile, the fourth lists the parameters that were tuned using Bayesian optimization, while the final lists the parameters that slice sampling was performed on.}}%
{\begin{tabular}{l c c c c}
     Model & Opt. objective & Params optimized & Bayes Opt. & Params slice sampled \\ \hline
     
    \textbf{MAP-$L$} & MAP & $\theta$, $\sigma^2$ & $\gamma$ & None \\
    \textbf{MAP-$L$ NL} & MAP & $\theta$, $\sigma^2$ &  $\gamma$ & $\alpha_W$, $\alpha_b$, $\sigma^2$ \\
    \textbf{Reg-$L$ NL} & Reg. ML & $\theta$, $\alpha_W$, $\alpha_b$, $\sigma^2$ & $\gamma_W$, $\gamma_b$ &  $\alpha_W$, $\alpha_b$, $\sigma^2$ \\
    \textbf{BN(ML)-$L$ NL} & ML & $\theta$, $\alpha_w$, $\alpha_b$, $a_0$, $b_0$ & None & $\alpha_W$, $\alpha_b$, $a_0$, $b_0$ \\
    \textbf{BN(BO)-$L$ NL} & ML & $\theta$ & $\alpha_W$, $\alpha_b$, $a_0$, $b_0$ &  $\alpha_W$, $\alpha_b$, $a_0$, $b_0$ \\

\end{tabular}}
\end{table}

\section{Additional Experimental Results} \label{apd:add_results}
In this appendix, we provide the full results from the main text, before briefly describing empirically the effect of slice sampling on the models.

\subsection{Tabulated Results and Average Rank Analysis} \label{apd:full_results}
On the next pages, we present tables of average test log likelihoods and test RMSEs for the UCI and UCI gap datasets for all models. For the UCI datasets, we present the average test log likelihoods and test RMSEs in \tableref{tab:UCI_LL,tab:UCI_RMSE} , as well as train log likelihoods and RMSEs in \tableref{tab:UCI_LL_train,tab:UCI_RMSE_train} . Following \citet{bui2016deep}, we also compute average ranks of the models across all splits of the standard UCI datasets. As in \citet{bui2016deep}, we additionally follow the procedure for the Friedman test as described in \citet{demvsar2006statistical}, generating the plots shown in \figureref{fig:LLs_rank,fig:RMSEs_rank}. These plots show the average rank of each method across all splits, where the difference between models is not statistically significant ($p < 0.05$) if the models are connected by a dark line, which is determined by the \textit{critical difference (CD)}.

\begin{figure}[htbp]
\floatconts
{fig:LLs_rank}
{\caption{Average ranks on the UCI datasets according to average test log likelihoods, generated as described in \citet{demvsar2006statistical}.}}
{\begin{tikzpicture}[scale=1]
\small
\draw (1, 0) -- coordinate (x axis mid) (10, 0);

\foreach \x in {1,...,10}
	\draw (\x, 0pt) -- (\x, -4pt) node[anchor=north] {\x};

\draw (3.29, 4pt) -- (3.29, 0.5) -- (0.8, 0.5) node[anchor=east] {BN(BO)-2 NL};
\draw (3.71, 4pt) -- (3.71, 0.85) -- (0.8, 0.85) node[anchor=east] {BN(ML)-2 NL};
\draw (4.55, 4pt) -- (4.55, 1.2) -- (0.8, 1.2) node[anchor=east] {Reg-2 NL};
\draw (5.41, 4pt) -- (5.41, 1.55) -- (0.8, 1.55) node[anchor=east] {BN(ML)-1 NL};
\draw (5.52, 4pt) -- (5.52, 1.9) -- (0.8, 1.9) node[anchor=east] {BN(BO)-1 NL};

\draw (8.28, 4pt) -- (8.28, 0.5) -- (10.2, 0.5) node[anchor=west] {MAP-1};
\draw (6.87, 4pt) -- (6.87, 0.85) -- (10.2, 0.85) node[anchor=west] {MAP-2};
\draw (6.27, 4pt) -- (6.27, 1.2) -- (10.2, 1.2) node[anchor=west] {MAP-1 NL};
\draw (5.57, 4pt) -- (5.57, 1.55) -- (10.2, 1.55) node[anchor=west] {Reg-1 NL};
\draw (5.53, 4pt) -- (5.53, 1.9) -- (10.2, 1.9) node[anchor=west] {MAP-2 NL};

\draw[ultra thick] (8, 2.5) -- (8.53, 2.5) node[above] {CD} -- (9.06, 2.5);

\draw[ultra thick] (3.24, 0.38) -- (3.76, 0.38);
\draw[ultra thick] (3.66, 0.73) -- (4.6, 0.73);
\draw[ultra thick] (4.5, 1.08) -- (5.62, 1.08);
\draw[ultra thick] (5.36, 0.73) -- (6.32, 0.73);
\draw[ultra thick] (6.22, 0.38) -- (6.92, 0.38);

\end{tikzpicture}}
\end{figure}
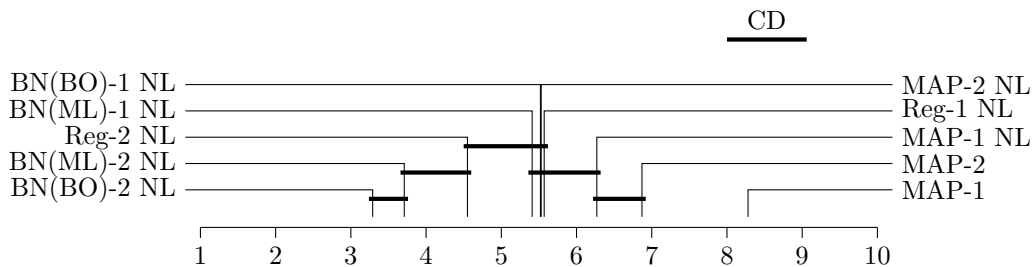

\begin{landscape}
\begin{table}
\tiny
\floatconts
  {tab:UCI_LL}%
  {\caption{Average Test Log Likelihoods (nats) on the UCI Datasets (Best Values in Bold, Errors are $\pm1$ Standard Error)}}%
{\begin{tabular}{l c c c c c c c c c c}
     \textbf{Dataset} & \textbf{MAP-1} & \textbf{MAP-2} & \textbf{MAP-1 NL} & \textbf{MAP-2 NL} & \textbf{Reg-1 NL} & \textbf{Reg-2 NL} & \textbf{BN(ML)-1 NL} & \textbf{BN(ML)-2 NL} & \textbf{BN(BO)-1 NL} & \textbf{BN(BO)-2 NL} \\ \hline
     
boston & -2.84 $\pm$ 0.15 & -2.85 $\pm$ 0.19 & -2.70 $\pm$ 0.11 & -2.71 $\pm$ 0.11 & -2.74 $\pm$ 0.11 & -2.80 $\pm$ 0.14 & -2.62 $\pm$ 0.09 & -2.59 $\pm$ 0.09 & $\mathbf{-2.58 \pm 0.09}$ & $\mathbf{-2.58 \pm 0.09}$ \\ 
concrete & -3.21 $\pm$ 0.04 & -3.20 $\pm$ 0.04 & -3.14 $\pm$ 0.04 & -3.20 $\pm$ 0.05 & -3.17 $\pm$ 0.05 & -3.21 $\pm$ 0.07 & -3.14 $\pm$ 0.06 & -3.15 $\pm$ 0.05 & -3.11 $\pm$ 0.05 & $\mathbf{-3.08 \pm 0.06}$ \\ 
energy & -0.87 $\pm$ 0.05 & -0.82 $\pm$ 0.09 & -0.65 $\pm$ 0.05 & -0.71 $\pm$ 0.09 & -0.74 $\pm$ 0.05 & -0.69 $\pm$ 0.06 & -0.66 $\pm$ 0.03 & -0.61 $\pm$ 0.05 & -0.69 $\pm$ 0.03 & $\mathbf{-0.54 \pm 0.04}$ \\ 
kin8nm & 1.07 $\pm$ 0.01 & 1.20 $\pm$ 0.01 & 1.12 $\pm$ 0.01 & 1.24 $\pm$ 0.01 & 1.15 $\pm$ 0.01 & 1.23 $\pm$ 0.01 & 1.12 $\pm$ 0.01 & 1.22 $\pm$ 0.01 & 1.12 $\pm$ 0.01 & $\mathbf{1.25 \pm 0.01}$ \\ 
naval & 5.19 $\pm$ 0.05 & 5.70 $\pm$ 0.09 & 6.31 $\pm$ 0.04 & 6.65 $\pm$ 0.04 & 8.24 $\pm$ 0.04 & 8.58 $\pm$ 0.03 & 7.85 $\pm$ 0.05 & $\mathbf{8.61 \pm 0.06}$ & 7.36 $\pm$ 0.08 & 7.96 $\pm$ 0.08 \\ 
power & -2.83 $\pm$ 0.01 & -2.80 $\pm$ 0.01 & -2.81 $\pm$ 0.01 & -2.78 $\pm$ 0.01 & -2.79 $\pm$ 0.01 & -2.76 $\pm$ 0.02 & -2.79 $\pm$ 0.01 & -2.75 $\pm$ 0.01 & -2.79 $\pm$ 0.01 & $\mathbf{-2.73 \pm 0.01}$ \\ 
protein & -2.96 $\pm$ 0.01 & -2.89 $\pm$ 0.00 & -2.94 $\pm$ 0.00 & -2.86 $\pm$ 0.00 & -2.87 $\pm$ 0.01 & -2.80 $\pm$ 0.01 & -2.86 $\pm$ 0.00 & -2.80 $\pm$ 0.01 & -2.87 $\pm$ 0.00 & $\mathbf{-2.78 \pm 0.01}$ \\ 
wine & -0.98 $\pm$ 0.02 & -0.97 $\pm$ 0.02 & -0.98 $\pm$ 0.01 & -0.98 $\pm$ 0.02 & -0.97 $\pm$ 0.02 & -0.98 $\pm$ 0.02 & -0.97 $\pm$ 0.01 & -0.97 $\pm$ 0.02 & $\mathbf{-0.96 \pm 0.01}$ & -0.98 $\pm$ 0.02 \\ 
yacht & -1.74 $\pm$ 0.25 & -2.03 $\pm$ 0.42 & -1.18 $\pm$ 0.17 & -1.75 $\pm$ 0.41 & -1.37 $\pm$ 0.20 & -1.29 $\pm$ 0.25 & -1.17 $\pm$ 0.07 & -1.13 $\pm$ 0.18 & -1.17 $\pm$ 0.07 & $\mathbf{-0.99 \pm 0.10}$ \\  \hline

    \textbf{Avg. Rank} & 8.28 $\pm$ 0.18 & 6.87 $\pm$ 0.19 & 6.27 $\pm$ 0.20 & 5.53 $\pm$ 0.19 & 5.57 $\pm$ 0.19 & 4.55 $\pm$ 0.24 & 5.41 $\pm$ 0.19 & 3.71 $\pm$ 0.19 & 5.52 $\pm$ 0.20 & $\mathbf{3.29 \pm 0.19}$
\end{tabular}}
\end{table}

\begin{table}
\tiny
\floatconts
  {tab:UCI_RMSE}%
  {\caption{Test RMSEs on the UCI Datasets (Best Values in Bold, Errors are $\pm 1$ Standard Error)}}%
{\begin{tabular}{l c c c c c c c c c c}
     \textbf{Dataset} & \textbf{MAP-1} & \textbf{MAP-2} & \textbf{MAP-1 NL} & \textbf{MAP-2 NL} & \textbf{Reg-1 NL} & \textbf{Reg-2 NL} & \textbf{BN(ML)-1 NL} & \textbf{BN(ML)-2 NL} & \textbf{BN(BO)-1 NL} & \textbf{BN(BO)-2 NL} \\ \hline
     
boston & 3.28 $\pm$ 0.23 & 3.25 $\pm$ 0.24 & 3.16 $\pm$ 0.19 & 3.17 $\pm$ 0.22 & 3.18 $\pm$ 0.22 & 3.20 $\pm$ 0.26 & 3.13 $\pm$ 0.21 & 3.05 $\pm$ 0.23 & 2.97 $\pm$ 0.19 & $\mathbf{2.91 \pm 0.19}$ \\ 
concrete & 5.41 $\pm$ 0.12 & 5.13 $\pm$ 0.12 & 5.14 $\pm$ 0.13 & 5.05 $\pm$ 0.11 & 5.03 $\pm$ 0.16 & 4.82 $\pm$ 0.14 & 5.08 $\pm$ 0.13 & 5.17 $\pm$ 0.12 & 4.96 $\pm$ 0.15 & $\mathbf{4.78 \pm 0.19}$ \\ 
energy & 0.52 $\pm$ 0.02 & 0.47 $\pm$ 0.02 & 0.44 $\pm$ 0.01 & 0.42 $\pm$ 0.02 & 0.46 $\pm$ 0.01 & 0.43 $\pm$ 0.02 & 0.46 $\pm$ 0.01 & 0.42 $\pm$ 0.01 & 0.48 $\pm$ 0.01 & $\mathbf{0.40 \pm 0.01}$ \\ 
kin8nm & 0.08 $\pm$ 0.00 & $\mathbf{0.07 \pm 0.00}$ & 0.08 $\pm$ 0.00 & $\mathbf{0.07 \pm 0.00}$ & 0.08 $\pm$ 0.00 & $\mathbf{0.07 \pm 0.00}$ & 0.08 $\pm$ 0.00 & $\mathbf{0.07 \pm 0.00}$ & 0.08 $\pm$ 0.00 & $\mathbf{0.07 \pm 0.00}$ \\ 
naval & $\mathbf{0.00 \pm 0.00}$ & $\mathbf{0.00 \pm 0.00}$ & $\mathbf{0.00 \pm 0.00}$ & $\mathbf{0.00 \pm 0.00}$ & $\mathbf{0.00 \pm 0.00}$ & $\mathbf{0.00 \pm 0.00}$ & $\mathbf{0.00 \pm 0.00}$ & $\mathbf{0.00 \pm 0.00}$ & $\mathbf{0.00 \pm 0.00}$ & $\mathbf{0.00 \pm 0.00}$ \\ 
power & 4.11 $\pm$ 0.04 & 3.99 $\pm$ 0.03 & 4.01 $\pm$ 0.04 & 3.90 $\pm$ 0.04 & 3.91 $\pm$ 0.04 & 3.74 $\pm$ 0.04 & 3.94 $\pm$ 0.04 & 3.73 $\pm$ 0.04 & 3.94 $\pm$ 0.04 & $\mathbf{3.70 \pm 0.04}$ \\ 
protein & 4.67 $\pm$ 0.03 & 4.33 $\pm$ 0.01 & 4.56 $\pm$ 0.01 & 4.24 $\pm$ 0.01 & 4.25 $\pm$ 0.02 & 3.94 $\pm$ 0.02 & 4.24 $\pm$ 0.01 & 3.94 $\pm$ 0.02 & 4.25 $\pm$ 0.01 & $\mathbf{3.88 \pm 0.02}$ \\ 
wine & 0.64 $\pm$ 0.01 & $\mathbf{0.63 \pm 0.01}$ & 0.64 $\pm$ 0.01 & $\mathbf{0.63 \pm 0.01}$ & 0.64 $\pm$ 0.01 & $\mathbf{0.63 \pm 0.01}$ & $\mathbf{0.63 \pm 0.01}$ & $\mathbf{0.63 \pm 0.01}$ & $\mathbf{0.63 \pm 0.01}$ & $\mathbf{0.63 \pm 0.01}$ \\ 
yacht & 0.73 $\pm$ 0.06 & 0.66 $\pm$ 0.06 & 0.61 $\pm$ 0.05 & 0.63 $\pm$ 0.05 & 0.64 $\pm$ 0.04 & 0.58 $\pm$ 0.06 & 0.79 $\pm$ 0.06 & $\mathbf{0.55 \pm 0.05}$ & 0.77 $\pm$ 0.06 & 0.66 $\pm$ 0.06 \\  \hline

    \textbf{Avg. Rank} & 8.36 $\pm$ 0.18 & 6.40 $\pm$ 0.20 & 6.73 $\pm$ 0.20 & 4.95 $\pm$ 0.18 & 5.47 $\pm$ 0.19 & 3.62 $\pm$ 0.21 & 6.13 $\pm$ 0.17 & 3.77 $\pm$ 0.18 & 6.11 $\pm$ 0.20 & $\mathbf{3.45 \pm 0.20}$
\end{tabular}}
\end{table}
\end{landscape}

\begin{landscape}
\begin{table}
\tiny
\floatconts
  {tab:UCI_LL_train}%
  {\caption{Average Train Log Likelihoods (nats) on the UCI Datasets (Errors are $\pm1$ Standard Error)}}%
{\begin{tabular}{l c c c c c c c c c c}
     \textbf{Dataset} & \textbf{MAP-1} & \textbf{MAP-2} & \textbf{MAP-1 NL} & \textbf{MAP-2 NL} & \textbf{Reg-1 NL} & \textbf{Reg-2 NL} & \textbf{BN(ML)-1 NL} & \textbf{BN(ML)-2 NL} & \textbf{BN(BO)-1 NL} & \textbf{BN(BO)-2 NL} \\ \hline
     
boston & -2.31 $\pm$ 0.05 & -2.25 $\pm$ 0.07 & -2.22 $\pm$ 0.05 & -2.14 $\pm$ 0.08 & -2.11 $\pm$ 0.09 & -2.00 $\pm$ 0.09 & -2.16 $\pm$ 0.06 & -2.20 $\pm$ 0.05 & -1.93 $\pm$ 0.04 & -1.77 $\pm$ 0.08 \\ 
concrete & -2.81 $\pm$ 0.03 & -2.65 $\pm$ 0.04 & -2.69 $\pm$ 0.03 & -2.55 $\pm$ 0.05 & -2.61 $\pm$ 0.04 & -2.21 $\pm$ 0.07 & -2.57 $\pm$ 0.04 & -2.62 $\pm$ 0.04 & -2.47 $\pm$ 0.03 & -2.15 $\pm$ 0.05 \\ 
energy & -0.49 $\pm$ 0.04 & -0.30 $\pm$ 0.05 & -0.20 $\pm$ 0.02 & -0.01 $\pm$ 0.05 & -0.29 $\pm$ 0.04 & 0.10 $\pm$ 0.06 & -0.33 $\pm$ 0.02 & 0.14 $\pm$ 0.04 & -0.43 $\pm$ 0.02 & 0.29 $\pm$ 0.06 \\ 
kin8nm & 1.15 $\pm$ 0.00 & 1.36 $\pm$ 0.01 & 1.22 $\pm$ 0.00 & 1.43 $\pm$ 0.01 & 1.29 $\pm$ 0.01 & 1.47 $\pm$ 0.01 & 1.25 $\pm$ 0.00 & 1.43 $\pm$ 0.01 & 1.25 $\pm$ 0.00 & 1.53 $\pm$ 0.01 \\ 
naval & 5.20 $\pm$ 0.05 & 5.70 $\pm$ 0.09 & 6.33 $\pm$ 0.04 & 6.68 $\pm$ 0.04 & 8.33 $\pm$ 0.04 & 8.68 $\pm$ 0.03 & 7.91 $\pm$ 0.05 & 8.67 $\pm$ 0.06 & 7.43 $\pm$ 0.08 & 8.03 $\pm$ 0.09 \\ 
power & -2.81 $\pm$ 0.00 & -2.77 $\pm$ 0.00 & -2.79 $\pm$ 0.00 & -2.74 $\pm$ 0.00 & -2.73 $\pm$ 0.00 & -2.60 $\pm$ 0.01 & -2.75 $\pm$ 0.00 & -2.60 $\pm$ 0.01 & -2.75 $\pm$ 0.00 & -2.55 $\pm$ 0.01 \\ 
protein & -2.95 $\pm$ 0.00 & -2.86 $\pm$ 0.00 & -2.92 $\pm$ 0.00 & -2.84 $\pm$ 0.00 & -2.82 $\pm$ 0.00 & -2.68 $\pm$ 0.01 & -2.83 $\pm$ 0.00 & -2.68 $\pm$ 0.01 & -2.82 $\pm$ 0.00 & -2.66 $\pm$ 0.01 \\ 
wine & -0.89 $\pm$ 0.01 & -0.86 $\pm$ 0.01 & -0.87 $\pm$ 0.01 & -0.83 $\pm$ 0.02 & -0.85 $\pm$ 0.01 & -0.86 $\pm$ 0.02 & -0.83 $\pm$ 0.01 & -0.83 $\pm$ 0.02 & -0.80 $\pm$ 0.01 & -0.82 $\pm$ 0.02 \\ 
yacht & -0.67 $\pm$ 0.13 & -0.32 $\pm$ 0.12 & 0.12 $\pm$ 0.07 & 0.49 $\pm$ 0.13 & -0.00 $\pm$ 0.09 & 0.93 $\pm$ 0.16 & -0.45 $\pm$ 0.06 & 0.76 $\pm$ 0.10 & -0.74 $\pm$ 0.05 & 0.43 $\pm$ 0.10 \\

\end{tabular}}
\end{table}

\begin{table}
\tiny
\floatconts
  {tab:UCI_RMSE_train}%
  {\caption{Train RMSEs on the UCI Datasets (Errors are $\pm1$ Standard Error)}}%
{\begin{tabular}{l c c c c c c c c c c}
     \textbf{Dataset} & \textbf{MAP-1} & \textbf{MAP-2} & \textbf{MAP-1 NL} & \textbf{MAP-2 NL} & \textbf{Reg-1 NL} & \textbf{Reg-2 NL} & \textbf{BN(ML)-1 NL} & \textbf{BN(ML)-2 NL} & \textbf{BN(BO)-1 NL} & \textbf{BN(BO)-2 NL} \\ \hline

boston & 2.51 $\pm$ 0.14 & 2.35 $\pm$ 0.18 & 2.29 $\pm$ 0.13 & 2.19 $\pm$ 0.18 & 2.15 $\pm$ 0.20 & 1.98 $\pm$ 0.22 & 2.20 $\pm$ 0.15 & 2.26 $\pm$ 0.13 & 1.69 $\pm$ 0.07 & 1.52 $\pm$ 0.13 \\ 
concrete & 4.03 $\pm$ 0.12 & 3.42 $\pm$ 0.16 & 3.61 $\pm$ 0.11 & 3.18 $\pm$ 0.17 & 3.35 $\pm$ 0.15 & 2.29 $\pm$ 0.14 & 3.20 $\pm$ 0.11 & 3.37 $\pm$ 0.12 & 2.86 $\pm$ 0.09 & 2.07 $\pm$ 0.10 \\ 
energy & 0.39 $\pm$ 0.02 & 0.32 $\pm$ 0.02 & 0.30 $\pm$ 0.01 & 0.25 $\pm$ 0.01 & 0.33 $\pm$ 0.01 & 0.23 $\pm$ 0.02 & 0.32 $\pm$ 0.01 & 0.20 $\pm$ 0.01 & 0.37 $\pm$ 0.01 & 0.18 $\pm$ 0.01 \\ 
kin8nm & 0.08 $\pm$ 0.00 & 0.06 $\pm$ 0.00 & 0.07 $\pm$ 0.00 & 0.06 $\pm$ 0.00 & 0.07 $\pm$ 0.00 & 0.06 $\pm$ 0.00 & 0.07 $\pm$ 0.00 & 0.06 $\pm$ 0.00 & 0.07 $\pm$ 0.00 & 0.05 $\pm$ 0.00 \\ 
naval & 0.00 $\pm$ 0.00 & 0.00 $\pm$ 0.00 & 0.00 $\pm$ 0.00 & 0.00 $\pm$ 0.00 & 0.00 $\pm$ 0.00 & 0.00 $\pm$ 0.00 & 0.00 $\pm$ 0.00 & 0.00 $\pm$ 0.00 & 0.00 $\pm$ 0.00 & 0.00 $\pm$ 0.00 \\ 
power & 4.03 $\pm$ 0.01 & 3.85 $\pm$ 0.01 & 3.92 $\pm$ 0.01 & 3.76 $\pm$ 0.01 & 3.72 $\pm$ 0.01 & 3.26 $\pm$ 0.04 & 3.78 $\pm$ 0.01 & 3.25 $\pm$ 0.03 & 3.79 $\pm$ 0.01 & 3.08 $\pm$ 0.03 \\ 
protein & 4.60 $\pm$ 0.02 & 4.23 $\pm$ 0.01 & 4.48 $\pm$ 0.01 & 4.15 $\pm$ 0.01 & 4.07 $\pm$ 0.01 & 3.52 $\pm$ 0.04 & 4.10 $\pm$ 0.00 & 3.52 $\pm$ 0.04 & 4.07 $\pm$ 0.01 & 3.45 $\pm$ 0.04 \\ 
wine & 0.59 $\pm$ 0.00 & 0.57 $\pm$ 0.01 & 0.58 $\pm$ 0.01 & 0.56 $\pm$ 0.01 & 0.57 $\pm$ 0.01 & 0.57 $\pm$ 0.01 & 0.56 $\pm$ 0.01 & 0.56 $\pm$ 0.01 & 0.54 $\pm$ 0.01 & 0.55 $\pm$ 0.01 \\ 
yacht & 0.42 $\pm$ 0.05 & 0.27 $\pm$ 0.05 & 0.22 $\pm$ 0.01 & 0.16 $\pm$ 0.03 & 0.27 $\pm$ 0.02 & 0.13 $\pm$ 0.02 & 0.36 $\pm$ 0.02 & 0.09 $\pm$ 0.02 & 0.49 $\pm$ 0.03 & 0.15 $\pm$ 0.02 \\

\end{tabular}}
\end{table}
\end{landscape}

\begin{figure}[htbp]
\floatconts
{fig:RMSEs_rank}
{\caption{Average ranks on the UCI datasets according to test RMSEs, generated as described in \cite{demvsar2006statistical}.}}
{\begin{tikzpicture}[scale=1]
\small
\draw (1, 0) -- coordinate (x axis mid) (10, 0);

\foreach \x in {1,...,10}
	\draw (\x, 0pt) -- (\x, -4pt) node[anchor=north] {\x};

\draw (3.45, 4pt) -- (3.45, 0.5) -- (0.8, 0.5) node[anchor=east] {BN(BO)-2 NL};
\draw (3.62, 4pt) -- (3.62, 0.85) -- (0.8, 0.85) node[anchor=east] {Reg-2 NL};
\draw (3.77, 4pt) -- (3.77, 1.2) -- (0.8, 1.2) node[anchor=east] {BN(ML)-2 NL};
\draw (4.95, 4pt) -- (4.95, 1.55) -- (0.8, 1.55) node[anchor=east] {MAP-2 NL};
\draw (5.47, 4pt) -- (5.47, 1.9) -- (0.8, 1.9) node[anchor=east] {Reg-1 NL};

\draw (8.36, 4pt) -- (8.36, 0.5) -- (10.2, 0.5) node[anchor=west] {MAP-1};
\draw (6.73, 4pt) -- (6.73, 0.85) -- (10.2, 0.85) node[anchor=west] {MAP-1 NL};
\draw (6.40, 4pt) -- (6.40, 1.2) -- (10.2, 1.2) node[anchor=west] {MAP-2};
\draw (6.13, 4pt) -- (6.13, 1.55) -- (10.2, 1.55) node[anchor=west] {BN(ML)-1 NL};
\draw (6.11, 4pt) -- (6.11, 1.9) -- (10.2, 1.9) node[anchor=west] {BN(BO)-1 NL};

\draw[ultra thick] (8, 2.5) -- (8.53, 2.5) node[above] {CD} -- (9.06, 2.5);

\draw[ultra thick] (3.4, 0.38) -- (3.82, 0.38);
\draw[ultra thick] (4.9, 0.38) -- (5.52, 0.38);
\draw[ultra thick] (5.42, 0.73) -- (6.45, 0.73);
\draw[ultra thick] (6.06, 0.38) -- (6.78, 0.38);

\end{tikzpicture}}
\end{figure}
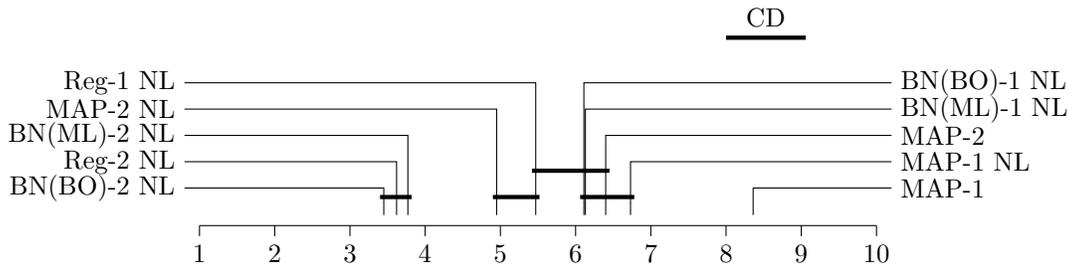

We make a few observations from these results. First, the two-layer marginal likelihood-based methods generally outperform the other methods, with the BN(BO)-2 NL model performing the best of all (although not significantly different from the BN(ML)-2 NL model according to the Friedman test). These are generally followed by the single-layer marginal-likelihood based methods, with the MAP-based methods performing the worst of all. This confirms our intuition that the more Bayesian versions of the models would yield better performance.

From the train log likelihoods and RMSEs, we observe that all the models exhibit overfitting for most of the datasets, with especially noticeable overfitting on `boston', `concrete', `energy', and `yacht'. The overfitting is generally worse on the two-layer models than the single-layer models, as there are more hyperparamters $\theta$ that can lead to overfitting in these models. Based off this trend, we expect that as the number of layers is increased further that the overfitting would worsen, thereby potentially limiting the use of neural linear models to smaller, shallower neural networks.

In \tableref{tab:UCI_gap_LL,tab:UCI_gap_RMSE} we show the test log likelihoods for the UCI gap datasets. As we are more concerned about whether the models can capture in-between uncertainty than the performance of the models on these datasets, we do not compute average ranks for these datasets. Additionally, since the test set is not within the same distribution as the training set, we do not show results on the training sets as they cannot be compared to the test performance. 

From these tables, we see that the MAP-$L$, MAP-$L$ NL, and Reg-$L$ NL models fail catastrophically on the `naval' and `energy' dataset. Additionally, MAP-1 performs especially poorly on `yacht', although it is not clear whether this can be termed `catastrophic'. While the BN NL models perform poorly on `naval' and `energy', by looking at the log likelihoods on the individual splits themselves we found that they were not actually failing catastrophically. This yet again confirms that the more fully Bayesian models are better, although it is surprising just how poorly the Reg-$L$ NL models perform. Additionally, because the overfitting we observed before worsens as the number of layers is increased, the performance of the BN NL models worsens as more layers are added.

\setlength{\tabcolsep}{4pt}
\begin{landscape}
\begin{table}
\tiny
\floatconts
  {tab:UCI_gap_LL}%
  {\caption{Average Test Log Likelihoods (nats) on the UCI Gap Datasets (Errors are $\pm1$ Standard Error)}}%
{\begin{tabular}{l c c c c c c c c c c}
     \textbf{Dataset} & \textbf{MAP-1} & \textbf{MAP-2} & \textbf{MAP-1 NL} & \textbf{MAP-2 NL} & \textbf{Reg-1 NL} & \textbf{Reg-2 NL} & \textbf{BN(ML)-1 NL} & \textbf{BN(ML)-2 NL} & \textbf{BN(BO)-1 NL} & \textbf{BN(BO)-2 NL} \\ \hline

boston & -2.79 $\pm$ 0.04 & -2.79 $\pm$ 0.08 & -2.72 $\pm$ 0.04 & -2.79 $\pm$ 0.09 & -2.86 $\pm$ 0.07 & -2.99 $\pm$ 0.17 & -2.76 $\pm$ 0.08 & -2.76 $\pm$ 0.06 & -2.81 $\pm$ 0.08 & -2.88 $\pm$ 0.08 \\ 
concrete & -3.61 $\pm$ 0.03 & -3.60 $\pm$ 0.06 & -3.61 $\pm$ 0.07 & -3.54 $\pm$ 0.05 & -4.70 $\pm$ 0.44 & -5.14 $\pm$ 0.27 & -3.74 $\pm$ 0.13 & -3.66 $\pm$ 0.13 & -3.72 $\pm$ 0.06 & -4.22 $\pm$ 0.17 \\ 
energy & -4.06 $\pm$ 1.27 & -6.55 $\pm$ 2.39 & -3.85 $\pm$ 1.20 & -5.89 $\pm$ 2.43 & -120.64 $\pm$ 78.92 & -95.79 $\pm$ 51.96 & -3.67 $\pm$ 1.11 & -4.22 $\pm$ 1.26 & -3.78 $\pm$ 1.16 & -3.87 $\pm$ 1.20 \\ 
kin8nm & 0.94 $\pm$ 0.05 & 1.09 $\pm$ 0.03 & 1.02 $\pm$ 0.06 & 1.15 $\pm$ 0.04 & 1.08 $\pm$ 0.04 & 1.18 $\pm$ 0.03 & 1.03 $\pm$ 0.03 & 1.15 $\pm$ 0.02 & 1.06 $\pm$ 0.02 & 1.18 $\pm$ 0.02 \\ 
naval & -12.06 $\pm$ 4.98 & -61.22 $\pm$ 28.17 & -7.43 $\pm$ 3.87 & -41.73 $\pm$ 14.80 & -716.29 $\pm$ 516.48 & -12294.53 $\pm$ 7478.34 & -4.30 $\pm$ 1.32 & -5.84 $\pm$ 1.40 & -1.17 $\pm$ 0.80 & -2.29 $\pm$ 0.99 \\ 
power & -2.87 $\pm$ 0.03 & -2.89 $\pm$ 0.04 & -2.87 $\pm$ 0.02 & -2.87 $\pm$ 0.02 & -3.19 $\pm$ 0.21 & -3.37 $\pm$ 0.22 & -2.93 $\pm$ 0.04 & -3.16 $\pm$ 0.08 & -2.96 $\pm$ 0.05 & -3.13 $\pm$ 0.09 \\ 
protein & -3.07 $\pm$ 0.01 & -3.06 $\pm$ 0.01 & -3.07 $\pm$ 0.01 & -3.06 $\pm$ 0.02 & -3.17 $\pm$ 0.04 & -3.29 $\pm$ 0.07 & -3.14 $\pm$ 0.03 & -3.22 $\pm$ 0.05 & -3.09 $\pm$ 0.03 & -3.22 $\pm$ 0.05 \\ 
wine & -0.96 $\pm$ 0.01 & -0.96 $\pm$ 0.01 & -0.96 $\pm$ 0.01 & -0.96 $\pm$ 0.01 & -1.03 $\pm$ 0.06 & -0.98 $\pm$ 0.01 & -0.96 $\pm$ 0.01 & -0.97 $\pm$ 0.01 & -0.98 $\pm$ 0.01 & -0.97 $\pm$ 0.02 \\ 
yacht & -3.69 $\pm$ 1.47 & -2.07 $\pm$ 0.30 & -2.58 $\pm$ 0.89 & -1.76 $\pm$ 0.27 & -2.20 $\pm$ 0.30 & -2.61 $\pm$ 0.57 & -1.81 $\pm$ 0.17 & -3.00 $\pm$ 0.77 & -1.72 $\pm$ 0.17 & -2.14 $\pm$ 0.28 \\ 

\end{tabular}}
\end{table}

\setlength{\tabcolsep}{6pt}
\begin{table}
\tiny
\floatconts
  {tab:UCI_gap_RMSE}%
  {\caption{Test RMSEs on the UCI Gap Datasets (Errors are $\pm1$ Standard Error)}}%
{\begin{tabular}{l c c c c c c c c c c}
     \textbf{Dataset} & \textbf{MAP-1} & \textbf{MAP-2} & \textbf{MAP-1 NL} & \textbf{MAP-2 NL} & \textbf{Reg-1 NL} & \textbf{Reg-2 NL} & \textbf{BN(ML)-1 NL} & \textbf{BN(ML)-2 NL} & \textbf{BN(BO)-1 NL} & \textbf{BN(BO)-2 NL} \\ \hline
    
boston & 3.82 $\pm$ 0.16 & 3.54 $\pm$ 0.14 & 3.71 $\pm$ 0.15 & 3.59 $\pm$ 0.14 & 3.69 $\pm$ 0.17 & 3.71 $\pm$ 0.21 & 3.67 $\pm$ 0.18 & 3.62 $\pm$ 0.17 & 3.85 $\pm$ 0.23 & 3.87 $\pm$ 0.21 \\ 
concrete & 7.79 $\pm$ 0.18 & 7.78 $\pm$ 0.23 & 7.68 $\pm$ 0.23 & 7.44 $\pm$ 0.17 & 8.21 $\pm$ 0.48 & 8.27 $\pm$ 0.39 & 7.69 $\pm$ 0.51 & 7.33 $\pm$ 0.36 & 7.74 $\pm$ 0.31 & 9.20 $\pm$ 0.55 \\ 
energy & 2.83 $\pm$ 0.99 & 3.70 $\pm$ 1.33 & 3.09 $\pm$ 1.17 & 3.48 $\pm$ 1.21 & 4.24 $\pm$ 2.11 & 3.83 $\pm$ 1.49 & 4.15 $\pm$ 1.64 & 4.10 $\pm$ 1.64 & 4.76 $\pm$ 1.98 & 4.58 $\pm$ 1.87 \\ 
kin8nm & 0.09 $\pm$ 0.01 & 0.08 $\pm$ 0.00 & 0.09 $\pm$ 0.01 & 0.07 $\pm$ 0.00 & 0.08 $\pm$ 0.00 & 0.07 $\pm$ 0.00 & 0.09 $\pm$ 0.00 & 0.08 $\pm$ 0.00 & 0.08 $\pm$ 0.00 & 0.07 $\pm$ 0.00 \\ 
naval & 0.02 $\pm$ 0.00 & 0.03 $\pm$ 0.00 & 0.02 $\pm$ 0.00 & 0.03 $\pm$ 0.00 & 0.01 $\pm$ 0.00 & 0.01 $\pm$ 0.00 & 0.01 $\pm$ 0.00 & 0.01 $\pm$ 0.00 & 0.01 $\pm$ 0.00 & 0.01 $\pm$ 0.00 \\ 
power & 4.24 $\pm$ 0.12 & 4.33 $\pm$ 0.18 & 4.25 $\pm$ 0.09 & 4.27 $\pm$ 0.08 & 5.17 $\pm$ 0.60 & 5.23 $\pm$ 0.43 & 4.49 $\pm$ 0.15 & 5.17 $\pm$ 0.28 & 4.66 $\pm$ 0.21 & 5.27 $\pm$ 0.36 \\ 
protein & 5.16 $\pm$ 0.04 & 5.07 $\pm$ 0.06 & 5.13 $\pm$ 0.05 & 5.08 $\pm$ 0.06 & 5.23 $\pm$ 0.12 & 5.33 $\pm$ 0.16 & 5.27 $\pm$ 0.12 & 5.37 $\pm$ 0.17 & 5.14 $\pm$ 0.10 & 5.46 $\pm$ 0.17 \\ 
wine & 0.63 $\pm$ 0.01 & 0.63 $\pm$ 0.01 & 0.63 $\pm$ 0.01 & 0.63 $\pm$ 0.01 & 0.66 $\pm$ 0.02 & 0.64 $\pm$ 0.01 & 0.63 $\pm$ 0.01 & 0.64 $\pm$ 0.01 & 0.65 $\pm$ 0.01 & 0.64 $\pm$ 0.01 \\ 
yacht & 1.31 $\pm$ 0.14 & 1.05 $\pm$ 0.09 & 1.28 $\pm$ 0.14 & 1.01 $\pm$ 0.09 & 1.24 $\pm$ 0.11 & 1.22 $\pm$ 0.13 & 1.15 $\pm$ 0.11 & 1.31 $\pm$ 0.16 & 1.37 $\pm$ 0.15 & 1.59 $\pm$ 0.23 \\

\end{tabular}}
\end{table}

\end{landscape}


\subsection{Additional Results on the Effect of Hyperparameter Tuning}
\label{apd:hyp_tune}

We describe the setup for our experiments on the effect of hyperparameter tuning as well as provide additional results not in the main text. We first describe the ``reasonable'' hyperparameter values that we selected:

\paragraph{MAP} For the MAP baseline, we select a batch size of 32. We set $\gamma = 0.5$, corresponding to a unity prior variance. We set the two learning rates to the ADAM default of 1e-3 \citep{kingma2014adam}. Finally, we allow for approximately 10000 gradient steps (we ensure that the last epoch is completed, so that there are at least 10000 gradient steps).

\paragraph{MAP NL} For the MAP neural linear model, we take the above optimal MAP network and obtain 200 slice samples of $\alpha_W$ (the output weight prior variance), $\alpha_b$ (the output bias prior variance), and $\sigma^2$ for Bayesian linear regression. We initialize $\alpha_W = 1/50$ and $\alpha_b = 1$, to match the scaling used in \citet{neal1995bayesian}.

\paragraph{Regularized NL} We set $\gamma_W = \gamma_b = 0.5$, the learning rates to 1e-3 and the number of epochs to 5000.

\paragraph{Bayesian noise NL (ML)} We initialize $a_0 = b_0 = \alpha_W = \alpha_b = 1$, the learning rate to 1e-3, and the number of epochs to 5000.

\paragraph{Bayesian noise NL (BO)} We use the same hyperparameter settings as above, although in this case $a_0, b_0, \alpha_W$, and $\alpha_b$ will remain fixed.\\

The log likelihoods and RMSEs for each split are then compared to those obtained when hyperparameter tuning is allowed. We show the average test log likelihoods and test RMSEs for the models without hyperparameter tuning in \tableref{tab:UCI_single_LL,tab:UCI_single_RMSE} for the UCI datasets and \tableref{tab:UCI_gap_single_LL,tab:UCI_gap_single_RMSE} for the UCI gap datasets. We also visualize the results for the gap datasets in \figureref{fig:single_diffs_UCI_gap}.

These results show that in general the hyperparameter tuning is of essential importance, particularly for the two-layer cases. As a whole, the results are significantly worse than with hyperparameter tuning, and in particular, for each of the two-layer methods, there is at least one dataset where the results are catastrophically bad compared to the models with hyperparameter tuning. Somewhat by contrast, however, while the results for the gap datasets are worse for the methods that did not fail catastrophically, they are not catastrophically worse. For the methods that were not able to represent in-between uncertainty, however, we find that they now fail catastrophically on even more datasets.

We now verify that the differences induced by the hyperparameter tuning are indeed statistically significant. In order to do so, we use the Wilcoxon signed-rank test \citep{wilcoxon1992individual} as described in \citet{demvsar2006statistical}. By comparing each tuned model to its non-tuned counterpart over all splits, we arrive at the table shown in \tableref{tab:Hyp_Effects}. This shows that the difference is indeed statistically significant ($p < 0.05$) on both the standard and gap datasets for the vast majority of models, measured both by log likelihoods and RMSEs.

\begin{figure}[htbp]
\floatconts
  {fig:single_diffs_UCI_gap}
  {\caption{Differences in average test log likelihoods (nats) and RMSEs given by hyperparameter tuning for the UCI gap datasets}}
  {\includegraphics[width=\linewidth]{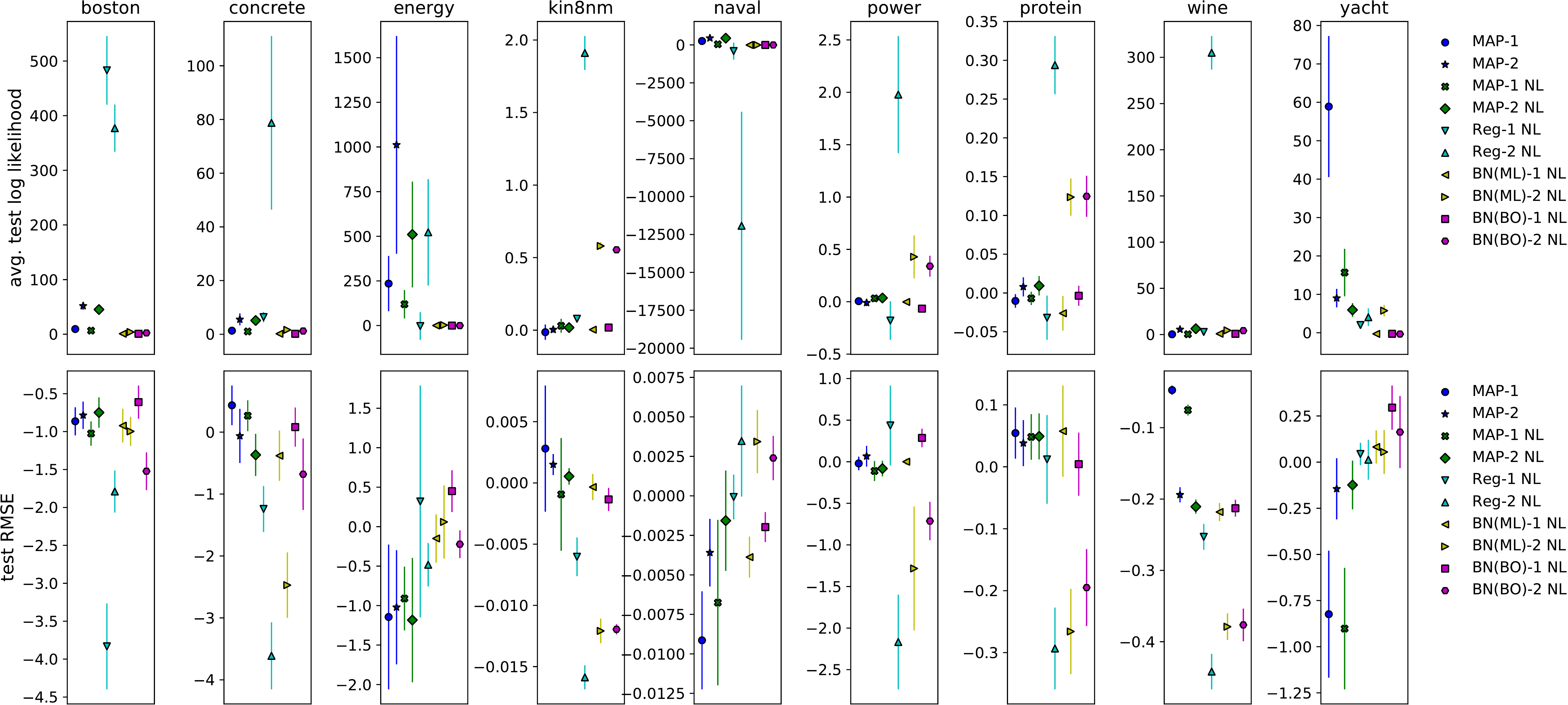}}
\end{figure}

\begin{table}
\small
\floatconts
{tab:Hyp_Effects}
{\caption{The Effect of Hyperparameter Tuning on the Models: (I) Improves, (W) Worsens, (N) Not Statistically Significant.}}
{\begin{tabular}{l c c c c}
\textbf{Model} & UCI: LL & UCI: RMSE & Gap: LL & Gap: RMSE \\ \hline
\textbf{MAP-1} & I & N & I & I\\
\textbf{MAP-2} & I & N & I & I\\
\textbf{MAP-1 NL} & I & I & I & I\\
\textbf{MAP-2 NL} & I & N & I & I\\
\textbf{Reg-1 NL} & I & I & I & I\\
\textbf{Reg-2 NL} & I & I & I & I\\
\textbf{BN(ML)-1 NL} & I & I & N & I\\
\textbf{BN(ML)-2 NL} & I & I & I & I\\
\textbf{BN(BO)-1 NL} & I & I & I & I\\
\textbf{BN(BO)-2 NL} & I & I & I & I \\
\end{tabular}}
\end{table}

\begin{landscape}
\begin{table}
\tiny
\floatconts
  {tab:UCI_single_LL}%
  {\caption{Average Test Log Likelihoods (nats) on the UCI Datasets Without Hyperparameter Tuning}}%
{\begin{tabular}{l c c c c c c c c c c}
     \textbf{Dataset} & \textbf{MAP-1} & \textbf{MAP-2} & \textbf{MAP-1 NL} & \textbf{MAP-2 NL} & \textbf{Reg-1 NL} & \textbf{Reg-2 NL} & \textbf{BN(ML)-1 NL} & \textbf{BN(ML)-2 NL} & \textbf{BN(BO)-1 NL} & \textbf{BN(BO)-2 NL} \\ \hline

boston & -4.45 $\pm$ 0.42 & -17.22 $\pm$ 1.56 & -4.25 $\pm$ 0.37 & -16.65 $\pm$ 1.78 & -259.51 $\pm$ 22.70 & -227.87 $\pm$ 36.75 & -3.53 $\pm$ 0.17 & -12.69 $\pm$ 0.61 & -3.63 $\pm$ 0.20 & -7.76 $\pm$ 0.32 \\ 
concrete & -3.19 $\pm$ 0.05 & -3.85 $\pm$ 0.15 & -3.16 $\pm$ 0.05 & -3.88 $\pm$ 0.16 & -3.83 $\pm$ 0.18 & -10.66 $\pm$ 1.10 & -3.11 $\pm$ 0.04 & -3.97 $\pm$ 0.19 & -3.12 $\pm$ 0.05 & -4.18 $\pm$ 0.15 \\ 
energy & -0.89 $\pm$ 0.08 & -0.99 $\pm$ 0.10 & -0.84 $\pm$ 0.06 & -0.88 $\pm$ 0.11 & -0.79 $\pm$ 0.08 & -1.91 $\pm$ 0.20 & -0.65 $\pm$ 0.03 & -0.98 $\pm$ 0.11 & -0.69 $\pm$ 0.03 & -0.55 $\pm$ 0.04 \\ 
kin8nm & 1.05 $\pm$ 0.01 & 1.18 $\pm$ 0.01 & 1.07 $\pm$ 0.01 & 1.22 $\pm$ 0.01 & 1.11 $\pm$ 0.01 & 0.48 $\pm$ 0.02 & 1.12 $\pm$ 0.01 & 0.79 $\pm$ 0.02 & 1.13 $\pm$ 0.01 & 0.78 $\pm$ 0.02 \\ 
naval & 4.96 $\pm$ 0.03 & 5.63 $\pm$ 0.03 & 6.12 $\pm$ 0.04 & 6.52 $\pm$ 0.03 & 7.84 $\pm$ 0.03 & 8.17 $\pm$ 0.02 & 7.44 $\pm$ 0.01 & 8.46 $\pm$ 0.03 & 7.35 $\pm$ 0.01 & 8.28 $\pm$ 0.01 \\ 
power & -2.83 $\pm$ 0.01 & -2.80 $\pm$ 0.01 & -2.81 $\pm$ 0.01 & -2.78 $\pm$ 0.01 & -2.79 $\pm$ 0.01 & -2.85 $\pm$ 0.02 & -2.79 $\pm$ 0.01 & -2.77 $\pm$ 0.02 & -2.79 $\pm$ 0.01 & -2.76 $\pm$ 0.02 \\ 
protein & -2.97 $\pm$ 0.01 & -2.91 $\pm$ 0.00 & -2.95 $\pm$ 0.00 & -2.90 $\pm$ 0.00 & -2.88 $\pm$ 0.00 & -2.81 $\pm$ 0.01 & -2.87 $\pm$ 0.00 & -2.81 $\pm$ 0.01 & -2.87 $\pm$ 0.00 & -2.81 $\pm$ 0.00 \\ 
wine & -1.05 $\pm$ 0.02 & -2.36 $\pm$ 0.10 & -1.08 $\pm$ 0.02 & -2.60 $\pm$ 0.10 & -1.76 $\pm$ 0.08 & -194.96 $\pm$ 13.60 & -1.34 $\pm$ 0.04 & -13.20 $\pm$ 0.25 & -1.48 $\pm$ 0.05 & -12.49 $\pm$ 0.21 \\ 
yacht & -4.31 $\pm$ 0.68 & -2.65 $\pm$ 0.59 & -3.25 $\pm$ 0.42 & -2.21 $\pm$ 0.49 & -1.36 $\pm$ 0.18 & -1.44 $\pm$ 0.24 & -1.16 $\pm$ 0.08 & -6.82 $\pm$ 1.15 & -1.13 $\pm$ 0.05 & -0.95 $\pm$ 0.10 \\ 

\end{tabular}}
\end{table}

\begin{table}
\tiny
\floatconts
  {tab:UCI_single_RMSE}%
  {\caption{Test RMSEs on the UCI Datasets Without Hyperparameter Tuning}}%
{\begin{tabular}{l c c c c c c c c c c}
     \textbf{Dataset} & \textbf{MAP-1} & \textbf{MAP-2} & \textbf{MAP-1 NL} & \textbf{MAP-2 NL} & \textbf{Reg-1 NL} & \textbf{Reg-2 NL} & \textbf{BN(ML)-1 NL} & \textbf{BN(ML)-2 NL} & \textbf{BN(BO)-1 NL} & \textbf{BN(BO)-2 NL} \\ \hline

boston & 3.15 $\pm$ 0.15 & 3.12 $\pm$ 0.14 & 3.21 $\pm$ 0.16 & 3.10 $\pm$ 0.15 & 5.21 $\pm$ 0.22 & 3.99 $\pm$ 0.28 & 3.23 $\pm$ 0.15 & 3.43 $\pm$ 0.18 & 3.43 $\pm$ 0.24 & 3.55 $\pm$ 0.15 \\ 
concrete & 4.97 $\pm$ 0.12 & 4.79 $\pm$ 0.14 & 4.93 $\pm$ 0.12 & 4.69 $\pm$ 0.15 & 4.83 $\pm$ 0.21 & 5.34 $\pm$ 0.22 & 4.82 $\pm$ 0.12 & 4.87 $\pm$ 0.24 & 4.78 $\pm$ 0.12 & 4.76 $\pm$ 0.12 \\ 
energy & 0.49 $\pm$ 0.02 & 0.43 $\pm$ 0.01 & 0.46 $\pm$ 0.01 & 0.40 $\pm$ 0.01 & 0.45 $\pm$ 0.02 & 0.42 $\pm$ 0.02 & 0.46 $\pm$ 0.01 & 0.42 $\pm$ 0.02 & 0.48 $\pm$ 0.02 & 0.40 $\pm$ 0.01 \\ 
kin8nm & 0.08 $\pm$ 0.00 & 0.07 $\pm$ 0.00 & 0.08 $\pm$ 0.00 & 0.07 $\pm$ 0.00 & 0.08 $\pm$ 0.00 & 0.08 $\pm$ 0.00 & 0.08 $\pm$ 0.00 & 0.08 $\pm$ 0.00 & 0.08 $\pm$ 0.00 & 0.08 $\pm$ 0.00 \\ 
naval & 0.00 $\pm$ 0.00 & 0.00 $\pm$ 0.00 & 0.00 $\pm$ 0.00 & 0.00 $\pm$ 0.00 & 0.00 $\pm$ 0.00 & 0.00 $\pm$ 0.00 & 0.00 $\pm$ 0.00 & 0.00 $\pm$ 0.00 & 0.00 $\pm$ 0.00 & 0.00 $\pm$ 0.00 \\ 
power & 4.10 $\pm$ 0.03 & 3.99 $\pm$ 0.04 & 4.02 $\pm$ 0.04 & 3.91 $\pm$ 0.03 & 3.94 $\pm$ 0.04 & 3.80 $\pm$ 0.04 & 3.94 $\pm$ 0.04 & 3.74 $\pm$ 0.04 & 3.94 $\pm$ 0.04 & 3.70 $\pm$ 0.04 \\ 
protein & 4.72 $\pm$ 0.02 & 4.45 $\pm$ 0.02 & 4.63 $\pm$ 0.01 & 4.39 $\pm$ 0.01 & 4.29 $\pm$ 0.01 & 3.95 $\pm$ 0.02 & 4.25 $\pm$ 0.02 & 3.95 $\pm$ 0.02 & 4.26 $\pm$ 0.01 & 3.96 $\pm$ 0.01 \\ 
wine & 0.64 $\pm$ 0.01 & 0.68 $\pm$ 0.01 & 0.66 $\pm$ 0.01 & 0.70 $\pm$ 0.01 & 0.74 $\pm$ 0.02 & 0.92 $\pm$ 0.02 & 0.69 $\pm$ 0.01 & 0.92 $\pm$ 0.02 & 0.73 $\pm$ 0.01 & 0.90 $\pm$ 0.02 \\ 
yacht & 0.66 $\pm$ 0.04 & 0.59 $\pm$ 0.05 & 0.63 $\pm$ 0.04 & 0.58 $\pm$ 0.05 & 0.60 $\pm$ 0.04 & 0.56 $\pm$ 0.05 & 0.77 $\pm$ 0.06 & 0.60 $\pm$ 0.06 & 0.75 $\pm$ 0.05 & 0.64 $\pm$ 0.07 \\ 

\end{tabular}}
\end{table}

\end{landscape}

\setlength{\tabcolsep}{4pt}
\begin{landscape}
\begin{table}
\tiny
\floatconts
  {tab:UCI_gap_single_LL}%
  {\caption{Average Test Log Likelihoods (nats) on the UCI Gap Datasets Without Hyperparameter Tuning}}%
{\begin{tabular}{l c c c c c c c c c c}
     \textbf{Dataset} & \textbf{MAP-1} & \textbf{MAP-2} & \textbf{MAP-1 NL} & \textbf{MAP-2 NL} & \textbf{Reg-1 NL} & \textbf{Reg-2 NL} & \textbf{BN(ML)-1 NL} & \textbf{BN(ML)-2 NL} & \textbf{BN(BO)-1 NL} & \textbf{BN(BO)-2 NL} \\ \hline

boston & -12.33 $\pm$ 1.48 & -54.66 $\pm$ 5.95 & -9.35 $\pm$ 1.02 & -48.01 $\pm$ 4.93 & -485.85 $\pm$ 61.72 & -380.12 $\pm$ 42.19 & -3.69 $\pm$ 0.14 & -6.82 $\pm$ 0.30 & -3.60 $\pm$ 0.11 & -5.29 $\pm$ 0.16 \\ 
concrete & -4.94 $\pm$ 0.31 & -9.13 $\pm$ 2.01 & -4.65 $\pm$ 0.27 & -8.68 $\pm$ 1.69 & -11.08 $\pm$ 1.75 & -83.89 $\pm$ 32.24 & -3.93 $\pm$ 0.07 & -5.31 $\pm$ 0.24 & -3.89 $\pm$ 0.11 & -5.46 $\pm$ 0.23 \\ 
energy & -238.97 $\pm$ 152.20 & -1018.49 $\pm$ 607.15 & -123.12 $\pm$ 76.99 & -516.00 $\pm$ 293.08 & -118.03 $\pm$ 67.94 & -618.22 $\pm$ 344.98 & -3.73 $\pm$ 1.19 & -6.62 $\pm$ 1.93 & -3.40 $\pm$ 1.02 & -3.81 $\pm$ 1.16 \\ 
kin8nm & 0.96 $\pm$ 0.03 & 1.09 $\pm$ 0.03 & 0.99 $\pm$ 0.02 & 1.13 $\pm$ 0.03 & 1.00 $\pm$ 0.03 & -0.73 $\pm$ 0.14 & 1.03 $\pm$ 0.02 & 0.57 $\pm$ 0.03 & 1.04 $\pm$ 0.02 & 0.62 $\pm$ 0.03 \\ 
naval & -271.04 $\pm$ 56.88 & -511.37 $\pm$ 97.14 & -53.08 $\pm$ 17.22 & -487.50 $\pm$ 97.54 & -309.49 $\pm$ 71.79 & -365.79 $\pm$ 100.30 & -2.01 $\pm$ 0.90 & -3.51 $\pm$ 1.15 & -1.48 $\pm$ 0.84 & -1.09 $\pm$ 0.90 \\ 
power & -2.87 $\pm$ 0.02 & -2.88 $\pm$ 0.03 & -2.90 $\pm$ 0.03 & -2.91 $\pm$ 0.04 & -3.01 $\pm$ 0.05 & -5.35 $\pm$ 0.76 & -2.92 $\pm$ 0.04 & -3.59 $\pm$ 0.27 & -2.90 $\pm$ 0.03 & -3.47 $\pm$ 0.15 \\ 
protein & -3.06 $\pm$ 0.02 & -3.06 $\pm$ 0.02 & -3.06 $\pm$ 0.02 & -3.07 $\pm$ 0.02 & -3.13 $\pm$ 0.03 & -3.58 $\pm$ 0.09 & -3.11 $\pm$ 0.03 & -3.34 $\pm$ 0.07 & -3.09 $\pm$ 0.03 & -3.35 $\pm$ 0.06 \\ 
wine & -1.20 $\pm$ 0.03 & -6.46 $\pm$ 0.42 & -1.26 $\pm$ 0.04 & -7.24 $\pm$ 0.53 & -3.78 $\pm$ 0.34 & -305.93 $\pm$ 17.49 & -1.78 $\pm$ 0.04 & -5.43 $\pm$ 0.06 & -1.75 $\pm$ 0.04 & -5.19 $\pm$ 0.08 \\ 
yacht & -62.59 $\pm$ 17.77 & -11.08 $\pm$ 2.36 & -18.25 $\pm$ 5.93 & -7.69 $\pm$ 1.77 & -4.19 $\pm$ 0.61 & -6.65 $\pm$ 2.20 & -1.51 $\pm$ 0.08 & -8.74 $\pm$ 1.42 & -1.44 $\pm$ 0.11 & -1.79 $\pm$ 0.11 \\

\end{tabular}}
\end{table}

\setlength{\tabcolsep}{6pt}
\begin{table}
\tiny
\floatconts
  {tab:UCI_gap_single_RMSE}%
  {\caption{Test RMSEs on the UCI Gap Datasets Without Hyperparameter Tuning}}%
{\begin{tabular}{l c c c c c c c c c c}
     \textbf{Dataset} & \textbf{MAP-1} & \textbf{MAP-2} & \textbf{MAP-1 NL} & \textbf{MAP-2 NL} & \textbf{Reg-1 NL} & \textbf{Reg-2 NL} & \textbf{BN(ML)-1 NL} & \textbf{BN(ML)-2 NL} & \textbf{BN(BO)-1 NL} & \textbf{BN(BO)-2 NL} \\ \hline
    
boston & 4.69 $\pm$ 0.21 & 4.33 $\pm$ 0.23 & 4.74 $\pm$ 0.21 & 4.33 $\pm$ 0.22 & 7.52 $\pm$ 0.67 & 5.50 $\pm$ 0.31 & 4.59 $\pm$ 0.29 & 4.62 $\pm$ 0.23 & 4.46 $\pm$ 0.28 & 5.39 $\pm$ 0.32 \\ 
concrete & 7.35 $\pm$ 0.33 & 7.84 $\pm$ 0.42 & 7.42 $\pm$ 0.33 & 7.81 $\pm$ 0.39 & 9.45 $\pm$ 0.51 & 11.88 $\pm$ 0.71 & 8.07 $\pm$ 0.21 & 9.80 $\pm$ 0.43 & 7.66 $\pm$ 0.38 & 9.88 $\pm$ 0.45 \\ 
energy & 3.97 $\pm$ 1.47 & 4.72 $\pm$ 1.96 & 4.00 $\pm$ 1.48 & 4.67 $\pm$ 1.93 & 3.92 $\pm$ 1.63 & 4.31 $\pm$ 1.61 & 4.30 $\pm$ 1.84 & 4.04 $\pm$ 1.46 & 4.31 $\pm$ 1.78 & 4.80 $\pm$ 2.02 \\ 
kin8nm & 0.09 $\pm$ 0.00 & 0.08 $\pm$ 0.00 & 0.09 $\pm$ 0.00 & 0.07 $\pm$ 0.00 & 0.09 $\pm$ 0.00 & 0.09 $\pm$ 0.00 & 0.09 $\pm$ 0.00 & 0.09 $\pm$ 0.00 & 0.08 $\pm$ 0.00 & 0.09 $\pm$ 0.00 \\ 
naval & 0.03 $\pm$ 0.00 & 0.03 $\pm$ 0.00 & 0.03 $\pm$ 0.01 & 0.03 $\pm$ 0.00 & 0.01 $\pm$ 0.00 & 0.01 $\pm$ 0.00 & 0.01 $\pm$ 0.00 & 0.01 $\pm$ 0.00 & 0.01 $\pm$ 0.00 & 0.01 $\pm$ 0.00 \\ 
power & 4.26 $\pm$ 0.07 & 4.27 $\pm$ 0.10 & 4.36 $\pm$ 0.12 & 4.35 $\pm$ 0.13 & 4.73 $\pm$ 0.18 & 7.40 $\pm$ 0.96 & 4.49 $\pm$ 0.17 & 6.45 $\pm$ 0.98 & 4.38 $\pm$ 0.13 & 5.98 $\pm$ 0.50 \\ 
protein & 5.10 $\pm$ 0.06 & 5.03 $\pm$ 0.07 & 5.08 $\pm$ 0.06 & 5.03 $\pm$ 0.08 & 5.22 $\pm$ 0.10 & 5.63 $\pm$ 0.17 & 5.21 $\pm$ 0.12 & 5.64 $\pm$ 0.22 & 5.14 $\pm$ 0.10 & 5.66 $\pm$ 0.18 \\ 
wine & 0.68 $\pm$ 0.01 & 0.82 $\pm$ 0.01 & 0.71 $\pm$ 0.01 & 0.84 $\pm$ 0.01 & 0.91 $\pm$ 0.03 & 1.08 $\pm$ 0.02 & 0.85 $\pm$ 0.01 & 1.01 $\pm$ 0.02 & 0.86 $\pm$ 0.01 & 1.02 $\pm$ 0.02 \\ 
yacht & 2.13 $\pm$ 0.34 & 1.20 $\pm$ 0.16 & 2.18 $\pm$ 0.34 & 1.13 $\pm$ 0.15 & 1.19 $\pm$ 0.11 & 1.21 $\pm$ 0.13 & 1.07 $\pm$ 0.06 & 1.25 $\pm$ 0.16 & 1.08 $\pm$ 0.13 & 1.43 $\pm$ 0.15 \\

\end{tabular}}
\end{table}

\end{landscape}

\subsubsection{Comparison to MFVI and MCD}
\label{apd:Ms}
To ensure that this worse behavior is not because all models require hyperparameter tuning to perform reasonably well, we now compare these results to results for mean field variational inference (MFVI) and Monte Carlo dropout (MCD) without hyperparameter tuning. We implement MFVI according to \citet{blundell2015weight} using the local reparameterization trick \citep{kingma2015variational}. We set a unity prior variance and use a step size of 1e-3, using ADAM \citep{kingma2014adam} with approximately 25000 gradient steps and a batch size of 32. We allow the gradients to be estimated using 10 samples from the approximate posterior at each step. For testing we use 100 samples from the approximate posterior.

For MCD, we follow the implementation in \citet{gal2016dropout}. We set the dropout rate to $p=0.05$ with weight decay corresponding to unity prior variance. We again use a learning rate of 1e-3, using ADAM \citep{kingma2014adam} with approximately 25000 gradient steps using a batch size of 32. For testing we use 100 samples generated by the neural network.

For the UCI datasets, we tabulate the test log likelihoods and test RMSEs for one- and two-layer architectures in \tableref{tab:UCI_Ms_LL,tab:UCI_Ms_RMSE}. These generally show reasonable values for each dataset despite the absence of hyperparameter tuning: there is no dataset for which either method can be said to do catastrophically badly. In fact, the results for MFVI are largely competitive with the best results we obtained for the neural linear models using hyperparameter tuning. This difference suggests that the reason hyperparameter tuning is important in the neural linear models is because it is necessary to carefully regularize the weights, whereas being approximately Bayesian over all of the weights is not as sensitive to the choice of hyperparameters. Although the train log likelihoods and RMSEs are not reported here, they confirm this intuition: the neural linear models still suffer from substantial overfitting, whereas the overfitting we observed for MFVI and MCD is far less.

\setlength{\tabcolsep}{2.5pt}
\begin{table}
\tiny
\floatconts
	{tab:UCI_Ms_LL}
	{\caption{Average Test Log Likelihoods (nats) on the UCI Datasets for MFVI and Monte Carlo Dropout}}
	{\begin{tabular}{l c c c c c c c c c}
	\textbf{Model} & boston & concrete & energy & kin8nm & naval & power & protein & wine & yacht \\ \hline 
	\textbf{MFVI-1} & -2.60 $\pm$ 0.06 & -3.09 $\pm$ 0.03 & -0.74 $\pm$ 0.02 & 1.11 $\pm$ 0.01 & 5.91 $\pm$ 0.04 & -2.82 $\pm$ 0.01 & -2.94 $\pm$ 0.00 & -0.97 $\pm$ 0.01 & -1.25 $\pm$ 0.13 \\ 
	\textbf{MFVI-2} & -2.82 $\pm$ 0.04 & -3.10 $\pm$ 0.02 & -0.77 $\pm$ 0.02 & 1.24 $\pm$ 0.01 & 5.99 $\pm$ 0.08 & -2.81 $\pm$ 0.01 & -2.87 $\pm$ 0.00 & -0.98 $\pm$ 0.01 & -1.15 $\pm$ 0.05 \\ 
\textbf{MCD-1} & -2.71 $\pm$ 0.11 & -3.33 $\pm$ 0.02 & -1.89 $\pm$ 0.03 & 0.67 $\pm$ 0.01 & 3.31 $\pm$ 0.01 & -2.98 $\pm$ 0.01 & -3.01 $\pm$ 0.00 & -0.97 $\pm$ 0.02 & -2.48 $\pm$ 0.10 \\ 
\textbf{MCD-2} & -2.70 $\pm$ 0.12 & -3.17 $\pm$ 0.03 & -1.34 $\pm$ 0.02 & 0.74 $\pm$ 0.01 & 3.91 $\pm$ 0.02 & -2.92 $\pm$ 0.01 & -2.95 $\pm$ 0.00 & -1.16 $\pm$ 0.04 & -2.88 $\pm$ 0.22 \\ 
	\end{tabular}}
	
\end{table}

\begin{table}
\tiny
\floatconts
	{tab:UCI_Ms_RMSE}
	{\caption{Test RMSEs on the UCI Datasets for MFVI and Monte Carlo Dropout}}
	{\begin{tabular}{l c c c c c c c c c}
	\textbf{Model} & boston & concrete & energy & kin8nm & naval & power & protein & wine & yacht \\ \hline 
\textbf{MFVI-1} & 3.19 $\pm$ 0.18 & 5.24 $\pm$ 0.11 & 0.50 $\pm$ 0.01 & 0.08 $\pm$ 0.00 & 0.00 $\pm$ 0.00 & 4.07 $\pm$ 0.04 & 4.61 $\pm$ 0.02 & 0.64 $\pm$ 0.01 & 0.73 $\pm$ 0.05 \\ 
\textbf{MFVI-2} & 3.73 $\pm$ 0.25 & 5.35 $\pm$ 0.10 & 0.51 $\pm$ 0.01 & 0.07 $\pm$ 0.00 & 0.00 $\pm$ 0.00 & 4.02 $\pm$ 0.03 & 4.29 $\pm$ 0.02 & 0.64 $\pm$ 0.01 & 0.76 $\pm$ 0.04 \\ 
\textbf{MCD-1} & 3.24 $\pm$ 0.22 & 6.65 $\pm$ 0.15 & 1.40 $\pm$ 0.07 & 0.12 $\pm$ 0.00 & 0.01 $\pm$ 0.00 & 4.68 $\pm$ 0.04 & 4.89 $\pm$ 0.02 & 0.63 $\pm$ 0.01 & 1.85 $\pm$ 0.34 \\ 
\textbf{MCD-2} & 3.06 $\pm$ 0.15 & 5.63 $\pm$ 0.16 & 0.82 $\pm$ 0.04 & 0.11 $\pm$ 0.00 & 0.00 $\pm$ 0.00 & 4.40 $\pm$ 0.04 & 4.55 $\pm$ 0.02 & 0.65 $\pm$ 0.01 & 3.16 $\pm$ 0.62 \\ 

	\end{tabular}}
	
\end{table}

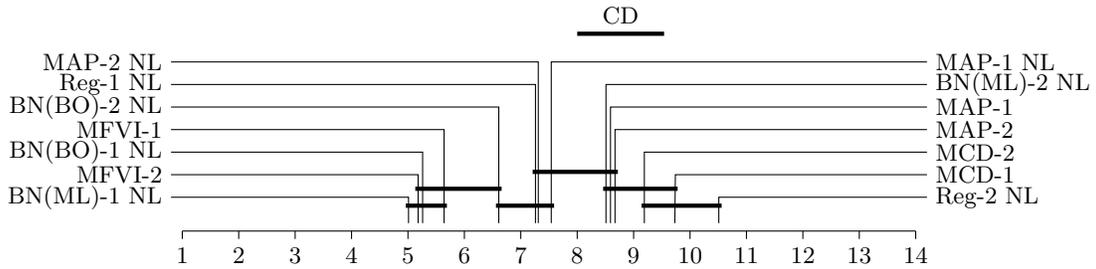
\begin{figure}[htbp]
\floatconts
{fig:LLs_rank_single}
{\caption{Average ranks of the single-run models on the UCI datasets according to average test log likelihoods, generated as described in \citet{demvsar2006statistical}.}}
{\begin{tikzpicture}[scale=0.75]
\footnotesize

\draw (1, 0) -- coordinate (x axis mid) (14, 0);

\foreach \x in {1,...,14}
	\draw (\x, 0pt) -- (\x, -4pt) node[anchor=north] {\x};

\draw (5.01, 4pt) -- (5.01, 0.6) -- (0.8, 0.6) node[anchor=east] {BN(ML)-1 NL};
\draw (5.18, 4pt) -- (5.18, 1) -- (0.8, 1) node[anchor=east] {MFVI-2};
\draw (5.26, 4pt) -- (5.26, 1.4) -- (0.8, 1.4) node[anchor=east] {BN(BO)-1 NL};
\draw (5.64, 4pt) -- (5.64, 1.8) -- (0.8, 1.8) node[anchor=east] {MFVI-1};
\draw (6.61, 4pt) -- (6.61, 2.2) -- (0.8, 2.2) node[anchor=east] {BN(BO)-2 NL};
\draw (7.26, 4pt) -- (7.26, 2.6) -- (0.8, 2.6) node[anchor=east] {Reg-1 NL};
\draw (7.31, 4pt) -- (7.31, 3) -- (0.8, 3) node[anchor=east] {MAP-2 NL};

\draw (7.54, 4pt) -- (7.54, 3) -- (14.2, 3) node[anchor=west] {MAP-1 NL};
\draw (8.51, 4pt) -- (8.51, 2.6) -- (14.2, 2.6) node[anchor=west] {BN(ML)-2 NL};
\draw (8.59, 4pt) -- (8.59, 2.2) -- (14.2, 2.2) node[anchor=west] {MAP-1};
\draw (8.67, 4pt) -- (8.67, 1.8) -- (14.2, 1.8) node[anchor=west] {MAP-2};
\draw (9.19, 4pt) -- (9.19, 1.4) -- (14.2, 1.4) node[anchor=west] {MCD-2};
\draw (9.73, 4pt) -- (9.74, 1) -- (14.2, 1) node[anchor=west] {MCD-1};
\draw (10.51, 4pt) -- (10.51, 0.6) -- (14.2, 0.6) node[anchor=west] {Reg-2 NL};

\draw[ultra thick] (8, 3.5) -- (8.77, 3.5) node[above] {CD} -- (9.54, 3.5);

\draw[ultra thick] (4.96, 0.45) -- (5.69, 0.45);
\draw[ultra thick] (5.13, 0.75) -- (6.66, 0.75);
\draw[ultra thick] (6.56, 0.45) -- (7.59, 0.45);
\draw[ultra thick] (7.21, 1.05) -- (8.72, 1.05);
\draw[ultra thick] (8.46, 0.75) -- (9.78, 0.75);
\draw[ultra thick] (9.14, 0.45) -- (10.56, 0.45); 

\end{tikzpicture}}
\end{figure}

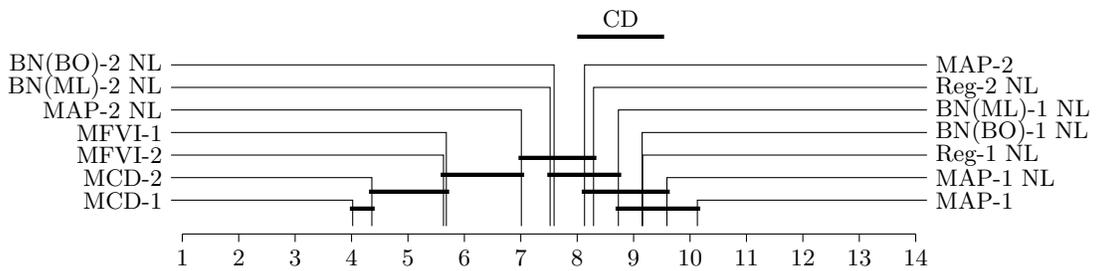
\begin{figure}[htbp]
\floatconts
{fig:RMSEs_rank_single}
{\caption{Average ranks of the single-run models on the UCI datasets according to test RMSEs, generated as described in \citet{demvsar2006statistical}.}}
{\begin{tikzpicture}[scale=0.75]
\footnotesize

\draw (1, 0) -- coordinate (x axis mid) (14, 0);

\foreach \x in {1,...,14}
	\draw (\x, 0pt) -- (\x, -4pt) node[anchor=north] {\x};

\draw (4.02, 4pt) -- (4.02, 0.6) -- (0.8, 0.6) node[anchor=east] {MCD-1};
\draw (4.36, 4pt) -- (4.36, 1) -- (0.8, 1) node[anchor=east] {MCD-2};
\draw (5.63, 4pt) -- (5.63, 1.4) -- (0.8, 1.4) node[anchor=east] {MFVI-2};
\draw (5.68, 4pt) -- (5.68, 1.8) -- (0.8, 1.8) node[anchor=east] {MFVI-1};
\draw (7.01, 4pt) -- (7.01, 2.2) -- (0.8, 2.2) node[anchor=east] {MAP-2 NL};
\draw (7.52, 4pt) -- (7.52, 2.6) -- (0.8, 2.6) node[anchor=east] {BN(ML)-2 NL};
\draw (7.59, 4pt) -- (7.59, 3) -- (0.8, 3) node[anchor=east] {BN(BO)-2 NL};

\draw (8.13, 4pt) -- (8.13, 3) -- (14.2, 3) node[anchor=west] {MAP-2};
\draw (8.29, 4pt) -- (8.29, 2.6) -- (14.2, 2.6) node[anchor=west] {Reg-2 NL};
\draw (8.73, 4pt) -- (8.73, 2.2) -- (14.2, 2.2) node[anchor=west] {BN(ML)-1 NL};
\draw (9.15, 4pt) -- (9.15, 1.8) -- (14.2, 1.8) node[anchor=west] {BN(BO)-1 NL};
\draw (9.16, 4pt) -- (9.16, 1.4) -- (14.2, 1.4) node[anchor=west] {Reg-1 NL};
\draw (9.59, 4pt) -- (9.59, 1) -- (14.2, 1) node[anchor=west] {MAP-1 NL};
\draw (10.13, 4pt) -- (10.13, 0.6) -- (14.2, 0.6) node[anchor=west] {MAP-1};

\draw[ultra thick] (8, 3.5) -- (8.77, 3.5) node[above] {CD} -- (9.54, 3.5);

\draw[ultra thick] (3.97, 0.45) -- (4.41, 0.45);
\draw[ultra thick] (4.31, 0.75) -- (5.73, 0.75);
\draw[ultra thick] (5.58, 1.05) -- (7.06, 1.05);
\draw[ultra thick] (6.96, 1.35) -- (8.34, 1.35);
\draw[ultra thick] (7.47, 1.05) -- (8.78, 1.05);
\draw[ultra thick] (8.08, 0.75) -- (9.64, 0.75);
\draw[ultra thick] (8.68, 0.45) -- (10.18, 0.45); 

\end{tikzpicture}}
\end{figure}

As with the results for the UCI datasets with tuned hyperparameters, we compute average ranks for the models across all splits and use the Friedman test as described in  \citet{demvsar2006statistical} to determine whether the differences are statistically significant. We plot the ranking using test log likelihoods in \figureref{fig:LLs_rank_single} and using test RMSEs in \figureref{fig:RMSEs_rank_single}. The rankings show that MCD performs poorly on average in terms of test log likelihood, whereas MFVI performs reasonably well for both log likelihood and RMSE. However, these rankings do not take into account that the neural linear models fail catastrophically on some, but not all, datasets since they only take the ordering of the methods into account and not how well they perform. Therefore, we still argue that for the standard UCI datasets MFVI and MCD are better since they perform relatively well across all datasets without the need for hyperparameter tuning.

Finally, we consider the performance of MFVI and MCD on the UCI gap datasets. We tabulate average test log likelihoods and test RMSEs in \tableref{tab:UCI_gap_Ms_LL,tab:UCI_gap_Ms_RMSE}. These echo the results in \citet{foong2019between}, showing catastrophic failure of MFVI to express `in-between' uncertainty for the `energy' and `naval' datasets. They also show that MCD fails catastrophically on the `energy' datasets, as suggested by theoretical results in \citet{foong2019pathologies}; however, we believe these are the first results in the literature that show catastrophic failure to express `in-between' uncertainty on a real dataset. The gap results therefore show one crucial advantage of the neural linear models over MFVI and MCD: their ability to express `in-between' uncertainty.

\setlength{\tabcolsep}{2pt}
\begin{table}
\tiny
\floatconts
	{tab:UCI_gap_Ms_LL}
	{\caption{Average Test Log Likelihoods (nats) on the UCI Gap Datasets for MFVI and Monte Carlo Dropout}}
	{\begin{tabular}{l c c c c c c c c c}
	\textbf{Model} & boston & concrete & energy & kin8nm & naval & power & protein & wine & yacht \\ \hline 
\textbf{MFVI-1} & -2.71 $\pm$ 0.06 & -3.53 $\pm$ 0.10 & -83.48 $\pm$ 53.70 & 1.05 $\pm$ 0.02 & -1318.01 $\pm$ 243.27 & -2.85 $\pm$ 0.02 & -3.04 $\pm$ 0.02 & -0.95 $\pm$ 0.01 & -1.84 $\pm$ 0.18 \\ 
\textbf{MFVI-2} & -2.88 $\pm$ 0.01 & -3.49 $\pm$ 0.07 & -11.93 $\pm$ 5.53 & 1.19 $\pm$ 0.02 & -1090.67 $\pm$ 192.18 & -2.94 $\pm$ 0.05 & -3.07 $\pm$ 0.03 & -0.97 $\pm$ 0.01 & -1.71 $\pm$ 0.18 \\ 
\textbf{MCD-1} & -2.88 $\pm$ 0.07 & -3.62 $\pm$ 0.04 & -20.71 $\pm$ 11.00 & 0.66 $\pm$ 0.02 & 2.46 $\pm$ 0.06 & -3.02 $\pm$ 0.03 & -3.08 $\pm$ 0.01 & -1.03 $\pm$ 0.02 & -2.67 $\pm$ 0.11 \\ 
\textbf{MCD-2} & -3.25 $\pm$ 0.12 & -3.89 $\pm$ 0.10 & -42.31 $\pm$ 23.94 & 0.73 $\pm$ 0.02 & -0.73 $\pm$ 0.55 & -3.23 $\pm$ 0.18 & -3.10 $\pm$ 0.02 & -1.52 $\pm$ 0.05 & -2.80 $\pm$ 0.16 \\ 

	\end{tabular}}
	
\end{table}

\setlength{\tabcolsep}{2.5pt}
\begin{table}
\tiny
\floatconts
	{tab:UCI_gap_Ms_RMSE}
	{\caption{Test RMSEs on the UCI Gap Datasets for MFVI and Monte Carlo Dropout}}
	{\begin{tabular}{l c c c c c c c c c}
	\textbf{Model} & boston & concrete & energy & kin8nm & naval & power & protein & wine & yacht \\ \hline 
\textbf{MFVI-1} & 3.78 $\pm$ 0.19 & 7.04 $\pm$ 0.33 & 4.30 $\pm$ 1.82 & 0.08 $\pm$ 0.00 & 0.03 $\pm$ 0.00 & 4.25 $\pm$ 0.12 & 5.02 $\pm$ 0.06 & 0.63 $\pm$ 0.01 & 1.30 $\pm$ 0.12 \\ 
\textbf{MFVI-2} & 3.70 $\pm$ 0.16 & 7.33 $\pm$ 0.25 & 2.58 $\pm$ 0.88 & 0.07 $\pm$ 0.00 & 0.03 $\pm$ 0.00 & 4.67 $\pm$ 0.23 & 5.05 $\pm$ 0.11 & 0.63 $\pm$ 0.01 & 1.26 $\pm$ 0.16 \\ 
\textbf{MCD-1} & 3.66 $\pm$ 0.12 & 8.00 $\pm$ 0.20 & 5.01 $\pm$ 1.72 & 0.12 $\pm$ 0.00 & 0.01 $\pm$ 0.00 & 4.87 $\pm$ 0.14 & 5.20 $\pm$ 0.05 & 0.65 $\pm$ 0.01 & 3.17 $\pm$ 0.55 \\ 
\textbf{MCD-2} & 3.58 $\pm$ 0.12 & 8.06 $\pm$ 0.24 & 5.18 $\pm$ 2.12 & 0.11 $\pm$ 0.00 & 0.02 $\pm$ 0.00 & 5.54 $\pm$ 0.64 & 5.18 $\pm$ 0.08 & 0.70 $\pm$ 0.01 & 3.75 $\pm$ 0.61 \\ 

	\end{tabular}}
\end{table}

\setlength{\tabcolsep}{6pt}

\subsection{Effect of Slice Sampling}
In this section, we briefly investigate the effect of slice sampling on the performance of the models. We first make plots of the predictive posterior distribution for each model trained on the toy problem of Section \ref{sec:Exp}. These plots are visible in \figureref{fig:no_slice_plots}. Note that the MAP-2 NL model simply becomes MAP inference. The most visible difference between \figureref{fig:no_slice_plots} and \figureref{fig:slice_plots} can be seen in the BN(BO)-2 NL model, which seems to have gained certainty at the edges while perhaps becoming slightly more uncertain in the gap; however, the effect in the gap is almost negligible. We observe the opposite effect in the MAP-2 NL model. Additionally, the Reg-2 NL model becomes slightly smoother. In general, however, it would seem that the effect of slice sampling for the toy problem is small.

\begin{figure}[htbp]
\floatconts
  {fig:no_slice_plots}
  {\caption{Predictive distributions for the toy problem, without slice sampling. Each shaded region represents one predictive standard deviation. Note that without slice sampling, the MAP-2 NL model simply becomes MAP-2.}}
  {\includegraphics[width=\linewidth]{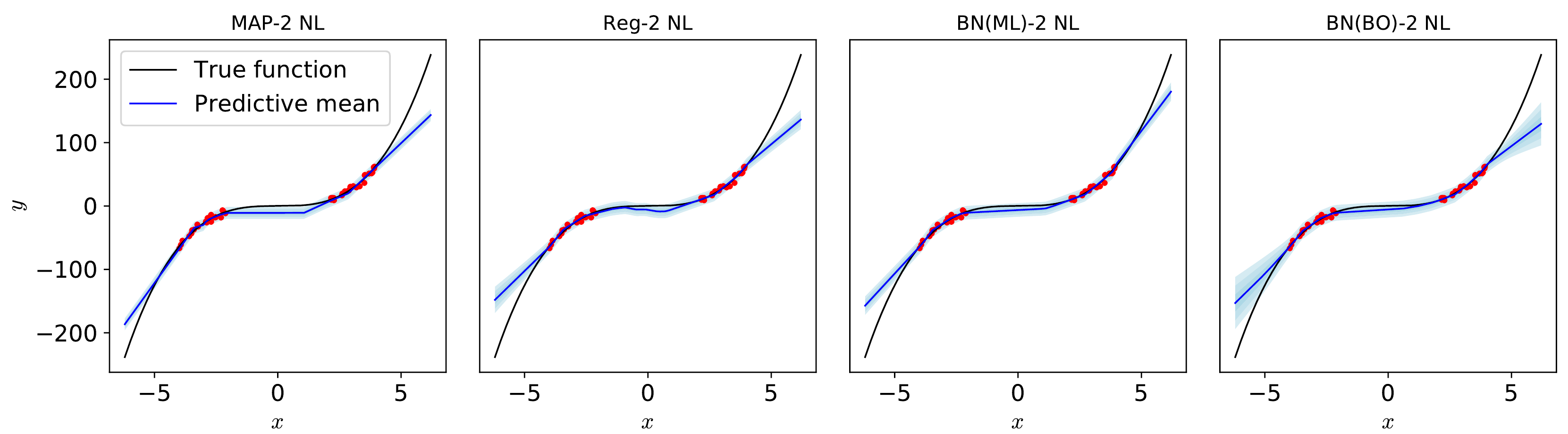}}
\end{figure}

We then plot the differences in the log likelihoods and RMSEs between the full models (with slice sampling) and the equivalent models without the final slice sampling step, to observe any quantitative differences. These plots are shown in \figureref{fig:slice_diffs_UCI} for the UCI datasets and \figureref{fig:slice_diffs_UCI_gap} for the UCI gap datasets. These plots do not give a clear picture of whether slice samping improves or worsens the performance of these models: it seems to depend on both the model and the dataset.

\begin{figure}[htbp]
\floatconts
  {fig:slice_diffs_UCI}
  {\caption{Differences in average test log likelihoods (nats) and RMSEs given by slice sampling for the UCI datasets}}
  {\includegraphics[width=0.9\linewidth]{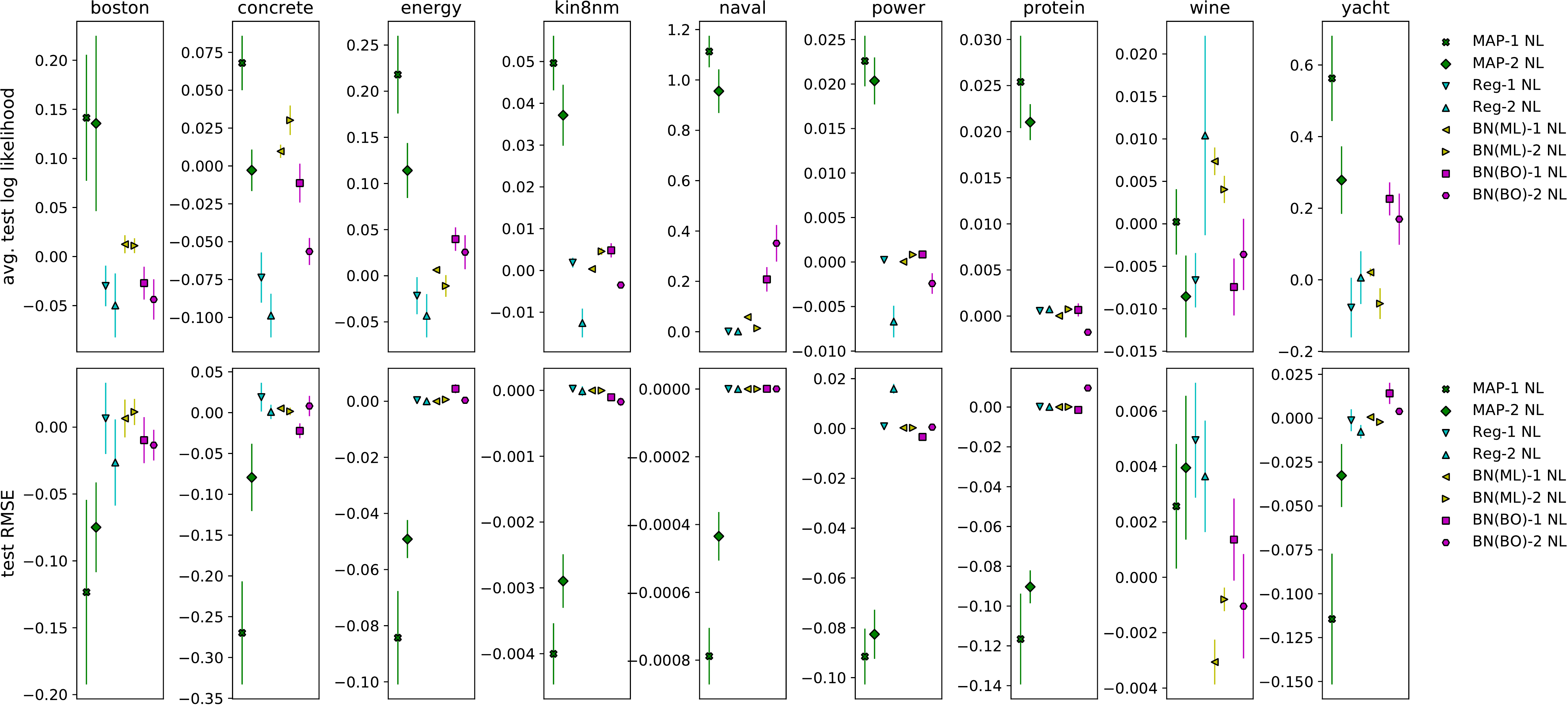}}
\end{figure}

\begin{figure}[htbp]
\floatconts
  {fig:slice_diffs_UCI_gap}
  {\caption{Differences in average test log likelihoods (nats) and RMSEs given by slice sampling for the UCI gap datasets}}
  {\includegraphics[width=0.9\linewidth]{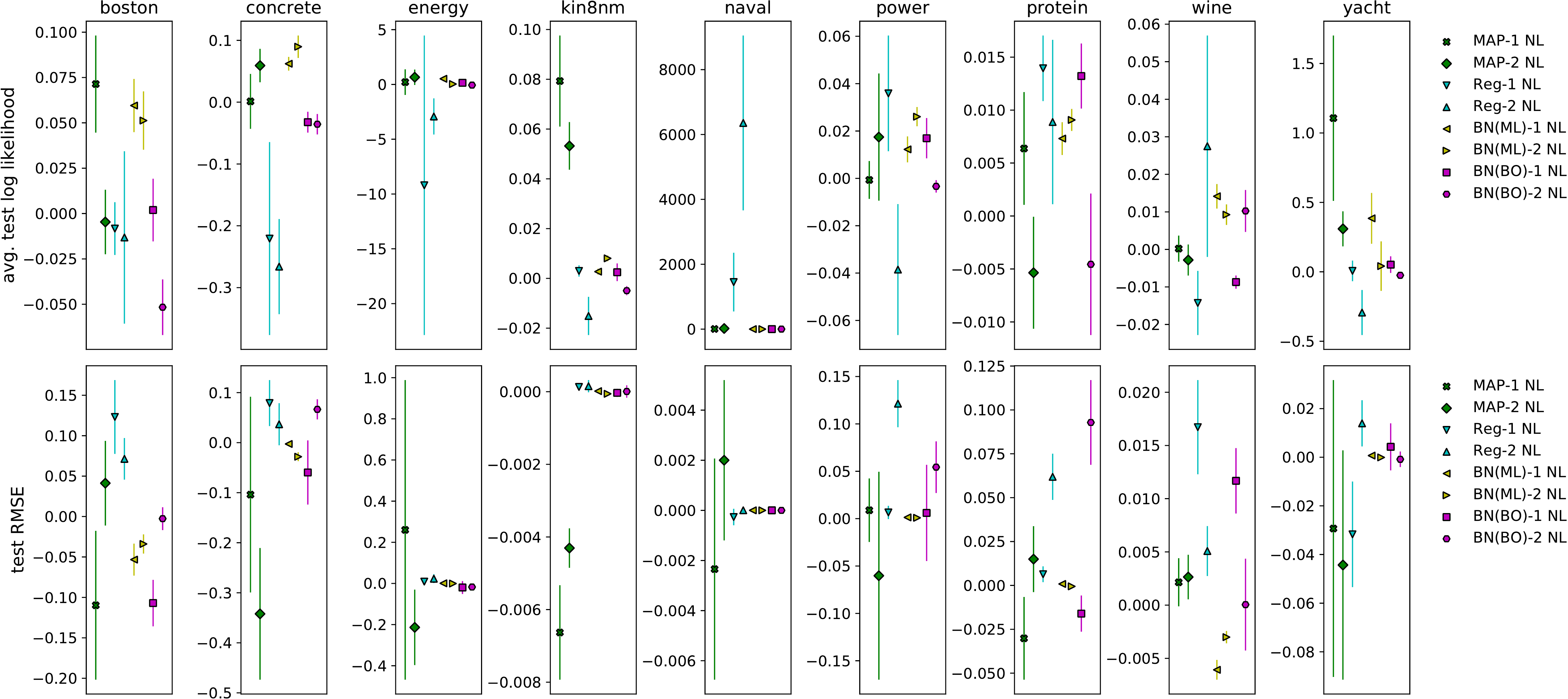}}
\end{figure}

To gain a clearer insight into whether slice sampling improves the performance of the neural linear models, we once again perform the Wilcoxon signed-rank test to compare the results obtained with slice sampling to those without. The results of this analysis is shown in \tableref{tab:Slice_Effects}. This shows that the majority of models are in fact improved by slice sampling, particularly when the improvement is measured in terms of the log likelihoods. However, in many cases, particularly when performance is measured in terms of RMSE, the effect of slice sampling is not statistically significant. Furthermore, it seems that performance for the Reg-$L$ NL and BN(BO)-2 NL models may be worsened by slice sampling.

In conclusion, in most cases performance will not be worsened by slice sampling. In particular, the MAP-$L$ NL models seem to benefit especially from slice sampling. However, slice sampling is likely detrimental to the Reg-$L$ NL models in all cases and potentially harmful for the BN(BO)-2 NL model when it comes to in-between uncertainty.

\begin{table}
\small
\floatconts
{tab:Slice_Effects}
{\caption{The Effect of Slice Sampling on the Models: (I) Improves, (W) Worsens, (N) Not Statistically Significant.}}
{\begin{tabular}{l c c c c}
\textbf{Model} & UCI: LL & UCI: RMSE & Gap: LL & Gap: RMSE \\ \hline
\textbf{MAP-1 NL} & I & I & I & I\\
\textbf{MAP-2 NL} & I & I & I & N\\
\textbf{Reg-1 NL} & W & N & I & W\\
\textbf{Reg-2 NL} & W & N & N & W\\
\textbf{BN(ML)-1 NL} & I & N & I & N\\
\textbf{BN(ML)-2 NL} & I & N & I & I\\
\textbf{BN(BO)-1 NL} & I & W & N & N\\
\textbf{BN(BO)-2 NL} & N & N & W & N \\
\end{tabular}}

\end{table}

\end{document}